%% file: paper.tex
\documentclass{article}
\usepackage[numbers,sort,compress]{natbib}
\usepackage[preprint]{neurips_2022}

\usepackage{times}
\usepackage{epsfig}
\usepackage{url}
\usepackage{xcolor}
\usepackage{booktabs}
\usepackage{tabularx}
\usepackage{multirow}
\usepackage{caption}
\usepackage{subcaption}
\usepackage{tikz}
\usepackage{amssymb}
\usepackage{amsmath}
\usepackage{dsfont}
\usepackage{bm}
\usepackage{wrapfig}
\usepackage{xspace}
\usepackage[final]{microtype}
\usepackage{enumitem}
\usepackage{adjustbox}

\input{preamble.tex}

\usepackage{etoolbox}
\usepackage{xspace}

\newcommand{\single}{\textsc{DistillNearest}\xspace}

\newcommand\multi[1]{%
  \ifstrempty{#1}{%
    \textsc{DistillWeighted}\xspace
  }{%
    \textsc{DistillWeighted}$(#1)$\xspace
  }%
}

\newcommand{\transfer}{\textsc{IN+Transfer}\xspace}
\newcommand{\fixmatch}{\textsc{IN+FixMatch}\xspace}
\newcommand{\randomSingle}{\textsc{DistillRandomSelection}\xspace}
\newcommand{\multiEqual}{\textsc{DistillEqual}\xspace}
\newcommand{\multiRandom}{\textsc{DistillRandomWeights}\xspace}

\newcommand\mypara[1]{\vspace{1.mm}\noindent\textbf{#1}}

\definecolor{orange_0}{rgb}{0.9937254901960785, 0.8501960784313726, 0.7043137254901961}
\definecolor{orange_1}{rgb}{0.9921568627450981, 0.6564705882352941, 0.3827450980392157}
\definecolor{orange_2}{rgb}{0.9545098039215686, 0.4399999999999999, 0.10666666666666655}
\definecolor{purple}{rgb}{0.5019607843137255, 0.0, 0.5019607843137255}
\definecolor{gradient_low}{rgb}{0.92907237, 0.68878959, 0.50411509}
\definecolor{gradient_medium}{rgb}{0.75861834, 0.25356035, 0.40663694}
\definecolor{gradient_high}{rgb}{0.29408557, 0.13721193, 0.38442775}
\newcommand{\gradientcolor}{\protect\tikz \protect\shade[left color=gradient_low, right color=gradient_high, middle color=gradient_medium] (0pt,0pt) rectangle +(20pt, 8pt);}

\definecolor{natural}{rgb}{0.7137,0.3333,0.3333}
\definecolor{specialized}{rgb}{0.4118,0.6431,0.4314}
\definecolor{structured}{rgb}{0.3254,0.4431,0.6666}
\definecolor{all}{rgb}{0.7529,0.4902,0.6471}

\newcommand{\vtabNatural}{\raisebox{0.5pt}{\tikz\fill[natural] (0,0) circle (.5ex);} \textit{Natural}\xspace}
\newcommand{\vtabSpecialized}{\raisebox{0.5pt}{\tikz\fill[specialized] (0,0) circle (.5ex);} \textit{Specialized}\xspace}
\newcommand{\vtabStructured}{\raisebox{0.5pt}{\tikz\fill[structured] (0,0) circle (.5ex);} \textit{Structured}\xspace}
\newcommand{\vtabMean}{\raisebox{0.5pt}{\tikz\fill[all] (0,0) circle (.5ex);} \textit{Mean}\xspace}

\newcommand{\eg}{\textit{e.g.}\xspace}

\usepackage[
    pagebackref=true,
    breaklinks=true,
    colorlinks,
    bookmarks=false
    ]{hyperref}

\title{Distilling from Similar Tasks for Transfer Learning on a Budget}

\author{%
  Kenneth Borup \\
  Aarhus University\\
  {\tt\small kennethborup@math.au.dk} \\
  \And
  Cheng Perng Phoo \\
  Cornell University \\
  {\tt\small cpphoo@cs.cornell.edu}
  \And
  Bharath Hariharan \\
  Cornell University \\
  {\tt\small bharathh@cs.cornell.edu}
}

\renewcommand{\L}{\mathcal{L}} 
\begin{document}
\maketitle

\input{0_abstract}

\input{1_intro}
\input{2_related_work}
\input{3_methodology}
\input{4_experiments}
\input{5_conclusion}

{\small
\setlength{\bibsep}{0pt}
\bibliographystyle{abbrvnat}
\bibliography{paper}
}

\input{6_supplementary_materials}

\end{document}

%% file: preamble.tex
\newcommand\eqdef{\mathrel{\overset{\makebox[0pt]{\mbox{\normalfont\tiny\sffamily def}}}{=}}} 
\newcommand{\trans}{{\intercal}}

\newcommand{\D}{\mathcal{D}}

\renewcommand{\L}{\mathcal{L}}
\newcommand{\M}{\mathcal{M}}

\newcommand{\R}{\mathbb{R}}
\renewcommand{\S}{\mathcal{S}}

\newcommand\bH{\mathbf{H}}
\newcommand\bI{\mathbf{I}}

\newcommand\bK{\mathbf{K}}
\newcommand\bL{\mathbf{L}}

\newcommand\bX{\mathbf{X}}
\newcommand\bY{\mathbf{Y}}

\newcommand\bx{\mathbf{x}}

\newcommand\by{\mathbf{y}}

%% file: 0_abstract.tex
\begin{abstract}
We address the challenge of getting efficient yet accurate recognition systems with limited labels. While recognition models improve with model size and amount of data, many specialized applications of computer vision have severe resource constraints both during training and inference. Transfer learning is an effective solution for training with few labels, however often at the expense of a computationally costly fine-tuning of large base models. We propose to mitigate this unpleasant trade-off between compute and accuracy via semi-supervised cross-domain distillation from a set of diverse source models. 
Initially, we show how to use task similarity metrics to select a single suitable source model to distill from, and that a good selection process is imperative for good downstream performance of a target model. We dub this approach \single. Though effective, \single assumes a single source model matches the target task, which is not always the case. To alleviate this, we propose a weighted multi-source distillation method to distill multiple source models trained on different domains weighted by their relevance for the target task into a single efficient model (named \multi{}). Our methods need no access to source data, and merely need features and pseudo-labels of the source models.
When the goal is accurate recognition under computational constraints, both \single{} and \multi{} approaches outperform both transfer learning from strong ImageNet initializations as well as state-of-the-art semi-supervised techniques such as FixMatch. Averaged over 8 diverse target tasks our multi-source method outperforms the baselines by 5.6\%-points and 4.5\%-points, respectively.
\end{abstract}

%% file: 1_intro.tex
\section{Introduction}

\begin{figure}[!ht]
    \begin{minipage}{0.48\textwidth}
        \centering
        \includegraphics[width=\linewidth]{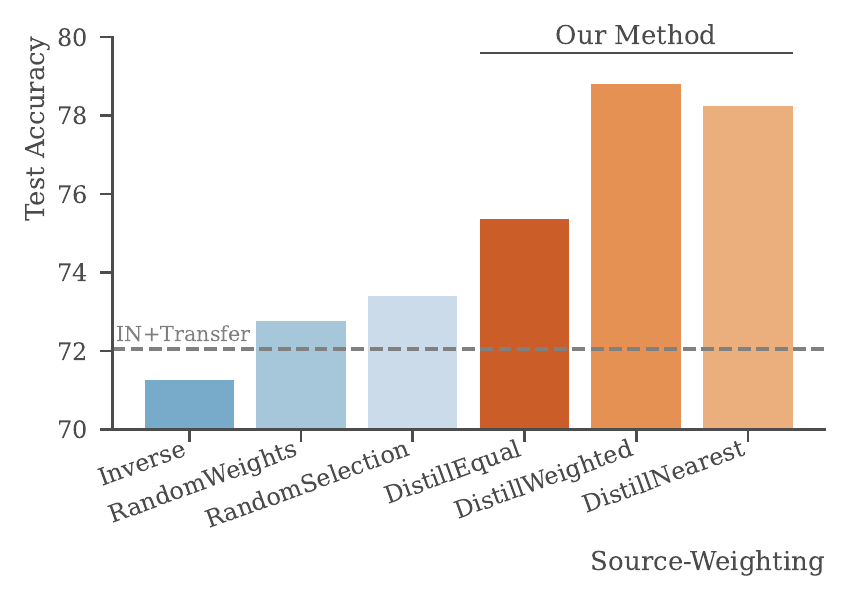}
        \caption{Average test accuracy over five target tasks with different methods for weighting source models for distillation. Our methods outperform the baselines and transfer learning from ImageNet. See Section \ref{sec: result_multi} for details.}
        \label{fig:alt_source_weights}
    \end{minipage}
    \hfill
    \begin{minipage}{0.48\textwidth}
        \centering
        \includegraphics[width=\linewidth]{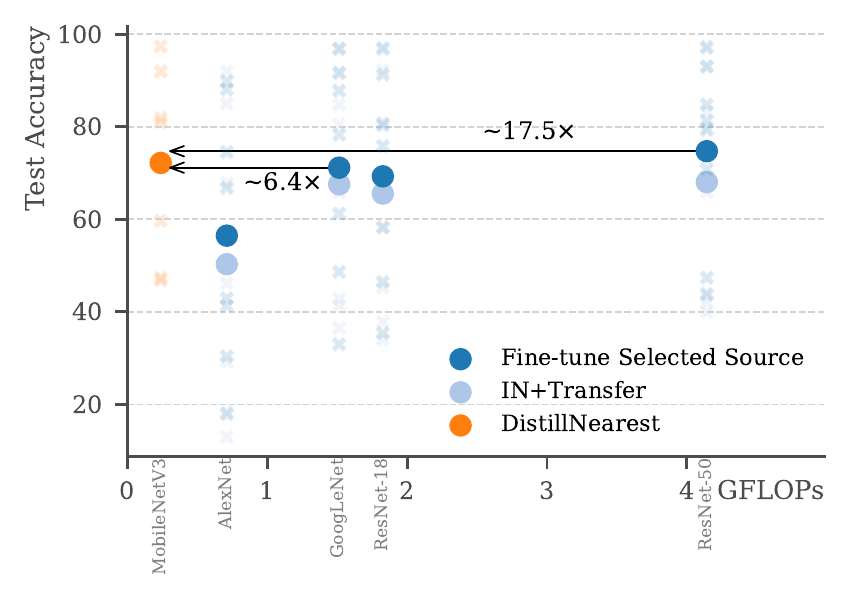}
        \caption{Average test accuracy over the 8 target tasks vs. compute requirements for a single forward pass at inference. Using \single{} with an efficient target architecture outperforms (or is comparable to) fine-tuning larger models.}
        \label{fig:acc_vs_compute}
    \end{minipage}
\end{figure}

Recognition models get more accurate the larger they are and the more data they are trained on \cite{sun2017revisiting, Zhai2022, Kolesnikov2020BigLearning}. This is a problem for many applications of interest in medicine (e.g. X-ray analysis) or science (e.g. satellite-image analysis) where both labeled training data, as well as computational resources needed to train such large models, are lacking.

The challenge of limited labeled data can potentially be alleviated by fine-tuning large-scale ``foundation models'' \cite{Zhai2022, Dehghani2023, Kolesnikov2020BigLearning}.
However, fine-tuning is computationally expensive, especially when one looks at foundation models with billions of parameters \cite{Dehghani2023}. Unfortunately, all evidence suggests that larger foundation models perform better at fine-tuning~\cite{Kolesnikov2020BigLearning, Zhai2022}.
This leaves downstream applications the unpleasant trade-off of expensive computational hardware for fine-tuning large models, or inaccurate results from smaller models. 
Motivated by this challenge, we ask \emph{can 
we train accurate models on tight data and compute budgets without fine-tuning large foundation models?}

To set the scene, we assume the existence of a diverse set (both in architecture and task) of pre-trained source models (or foundation models). 
We do not have the resources to fine-tune these models, but we assume we can perform inference on these models and extract features, \eg through APIs on cloud services \cite{bisong2019google, rekognition}.
For the target task, we assume that labeled data is very limited, but unlabeled data is available.
We then propose a simple and effective strategy for building an accurate model for the target task: \single.
Concretely, we first compute a measure of ``task similarity'' between our target task and each source model and rank the source models accordingly. Then we pseudo-label the unlabeled data using the most similar source model. These pseudo-labels  may not even be in the same label space as the target task, but we conjecture that due to the similarity between the source and target tasks, the pseudo-labels will still \emph{group} the target data points in a task-relevant manner. 
Finally, we train the target model using the pseudo-labels and the available ground truth labeled data. This allows us to bypass the large computations required to fine-tune source models and directly work on the target model. At the same time, we get to effectively use the knowledge of the large source model even if it is trained on a different task.

\single assumes that a \textit{single} best source model exists.
But for some target tasks, we might need to combine multiple source models to achieve a sufficiently diverse representation to distill. We, therefore, propose an extension of our approach that distills \emph{multiple (diverse) source models} trained on different domains, weighted by their relevance for the target task.
This extension obtains even further improvements on our target performance (see Figure \ref{fig:alt_source_weights}). We dub this method \multi{}.

\mypara{We summarize our contributions as follows:}
\begin{itemize}[leftmargin=9pt, itemsep=0pt, topsep=3pt]
    \item We train more than 200 models across a diverse set of source and target tasks using single-source distillation, and extensively show that the choice of source model is imperative for the predictive performance of the target model. To the best of our knowledge, no previous work has addressed how to efficiently select a teacher model for (cross-domain) distillation.
    \item We find that \textit{task similarity metrics} correlate well with predictive performance and can be used to efficiently select and weight source models for single- and multi-source distillation without access to any source data.
    \item We show that our approaches yield the best accuracy on multiple target tasks under compute and data constraints. We compare our \single{} and \multi{} methods to two baselines (transfer learning and FixMatch), as well as the naïve case of \multi{} with \emph{equal} weighting (called \multiEqual{}), among others. Averaged over 8 diverse datasets, our \multi{} outperforms the baselines with at least 4.5\% and in particular 17.5\% on CUB200.
\end{itemize}

\begin{figure*}[t]
    \centering
    \includegraphics[width=\linewidth]{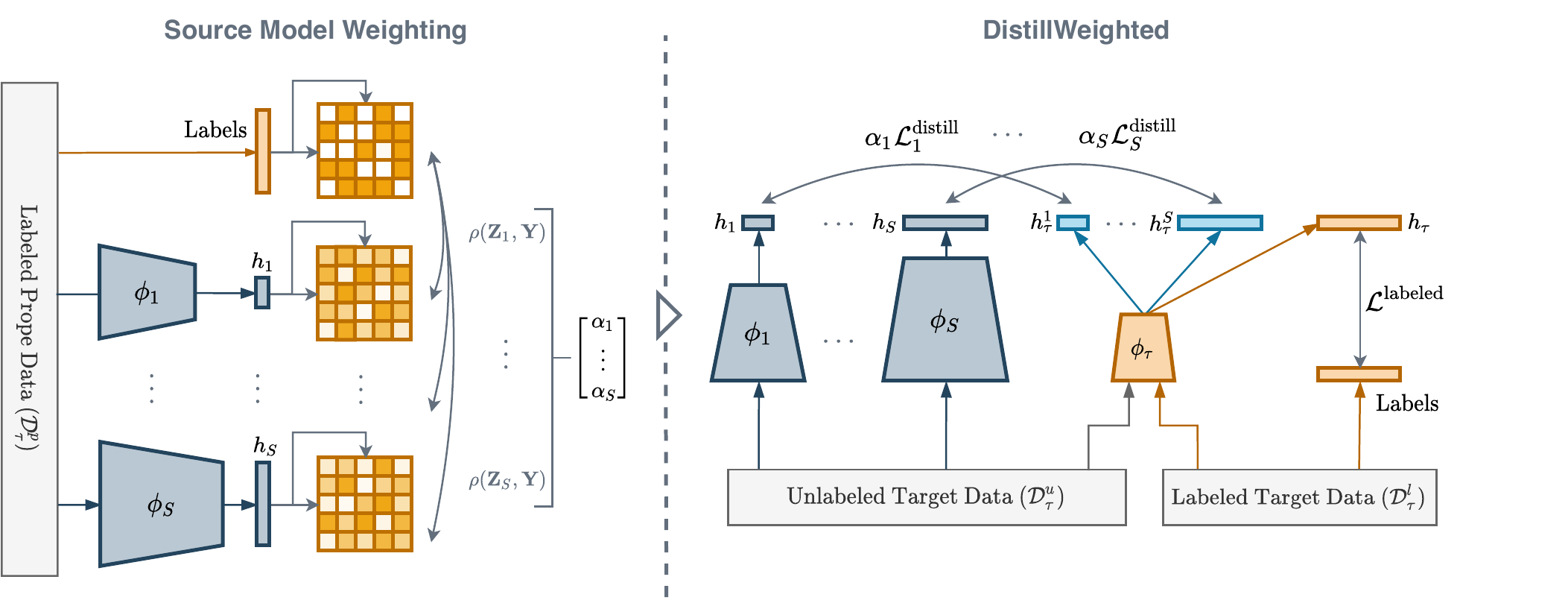} 
    \caption{We propose to weigh a set of $S$ source models, $\M_s = h_s \circ \phi_s$, by using task similarity metrics to estimate the alignment of each source model with the particular target task using a small probe set of labeled data, $\D_{\tau}^p$. Since the task similarity metrics are independent of feature dimension, we can utilize source models of any architecture and from any source task. We show that by choosing the weighting, $\alpha_1, \dots, \alpha_S$, this way we are able to improve performance over transfer from ImageNet and training with FixMatch amongst others (see \eg Table \ref{tbl:vs_finetuned_source_models} and Figure \ref{fig:natural_images_overall}).}
    \label{fig:method_figure}
\end{figure*}

%% file: 2_related_work.tex
\section{Related Work}
\mypara{Knowledge Distillation} One key aspect of our problem is to figure out how to compress single or multiple large foundation models into an efficient target model. A common approach is knowledge distillation \citep{ba2013deep, hinton2015distilling} where an efficient student model is trained to mimic the output of a larger teacher model. However, most single-teacher \citep{adriana2015fitnets, mirzadeh2019improved, park2019relational, Cho2019OnDistillation, borup2021even} or multi-teacher knowledge distillation \citep{you2017learning, fukuda2017efficient, tan2019multilingual, liu2020adaptive} research focuses on the closed set setup, where the teacher(s) and the student both attempts to tackle the same task. To the best of our knowledge, compressing models specializing in various tasks different from the target task has rarely been explored in the literature. Our paper explores this setup and illustrates that carefully distilling source models trained on different tasks can bring forth efficient yet accurate models. 

\mypara{Semi-Supervised Learning and Transfer}
Given our target tasks are specified in a semi-supervised setting, it is customary to review methods for semi-supervised learning (SSL). The key to SSL approaches is how to effectively propagate label information from a small labeled dataset to a large unlabeled dataset. Along this vein, methods such as pseudo-labeling/self-training \citep{lee2013pseudo, xie2020self} or consistency regularization \citep{tarvainen2017mean, berthelot2019mixmatch, sohn2020fixmatch} have shown remarkable results in reducing deep networks dependencies on large labeled datasets via unlabeled data. However, most SSL approaches focus on training models from scratch without considering the availability of pre-trained models. Given the increasing availability of large pre-trained models \citep{paszke2019automatic, wolf2019huggingface}, recent work has started exploring the intersection between transfer learning and SSL \citep{phoo2021selftraining, islam2021dynamic, abuduweili2021adaptive}. However, most of these works focus on how to transfer from a single pre-trained model to the target task. Our paper, however, explores an even more practical setup: how to transfer from multiple pre-trained models to a downstream task where in-domain unlabeled data are available. In principle, we could combine our approach with a lot of previous work on SSL to (potentially) gain even larger improvements, but to keep our method simple we leave such exploration to future work and focus on how to better utilize an available set of pre-trained models.

\mypara{Multi-Source Domain Adaptation}
Our setup also bears a resemblance with multi-source domain adaptation (MSDA) \citep{peng2019moment} in which the goal is to create a target model by leveraging multiple source models. However, MSDA methods often assume the source and target models share the same label space to perform domain alignment. We do not make such an assumption and in fact, focus on the case where the label space of source and target tasks have minimal to no overlap. Besides, a lot of the MSDA approaches \citep{zhao2018adversarial, xu2018deep, peng2019moment, zhao2020multi} rely on the availability of source data or the fact that the source and target tasks share the same model architecture to build domain invariant features. Given the discrepancy in assumptions between MSDA and our setup, we do not consider any methods from this line of work as baselines. 

\mypara{Transfer Learning From Multiple Sources}
Transfer learning from multiple different pre-trained models has been explored in different setups. \citet{Bolya2021ScalableLearning} focuses on how to select a single good pre-trained model to use as a model initialization whereas we explore how to distill an efficient model from the pre-trained models (i.e. our target architecture could be different from those of the source models). \citet{agostinelli2022transferability} focuses on how to select a subset of pre-trained models to construct an (fine-tuned) ensemble, whereas we focus on creating a single model. 
\citet{Li2021RepresentationStudents} focuses on creating a generalist representation by equally distilling multiple pre-trained models using proxy/source data (which often requires high-capacity models) whereas our goal is to construct an efficient specialist model using the target data. All these works have indicated the importance of exploring how to best leverage a large collection of pre-trained models but due to differences in setup and assumptions, we do not (and could not) compare to them. 

\mypara{Task Similarity / Transferability Metrics}
A key insight of our approach is to leverage the similarity between the target and source tasks to compare and weigh different pre-trained source models during distillation. Characterizing tasks (or similarities between tasks) is an open research question with various successes. A common approach is to embed tasks into a common vector space and characterize similarities in said space. Representative research along this line of work include \citet{achille2019task2vec,peng2020domain2vec,wallace2021can}. Another related line of work investigates transferability metrics \citep{tran2019transferability, bao2019information, nguyen2020leep, dwivedi2020duality, Dwivedi2019RepresentationLearning, Bolya2021ScalableLearning}. After all, one of the biggest use cases of task similarities is to predict how well a model transfers to new tasks.
Since it is not our intention to define new task similarity/transferability metrics for distillation, we use already established metrics that capture the similarity between source representations and one-hot labels to weigh the source models.
Under this purview, metrics that characterize similarities between features such as CKA \citep{Cortes2012AlgorithmsAlignment, kornblith2019similarity} and transferability metrics based on features \citep{Dwivedi2019RepresentationLearning, Bolya2021ScalableLearning} suffice.

%% file: 3_methodology.tex
\section{Problem Setting}
\begin{figure*}[!t]
    \begin{adjustbox}{width=1.2\textwidth,center}
        \centering
        \includegraphics[width=1.2\linewidth]{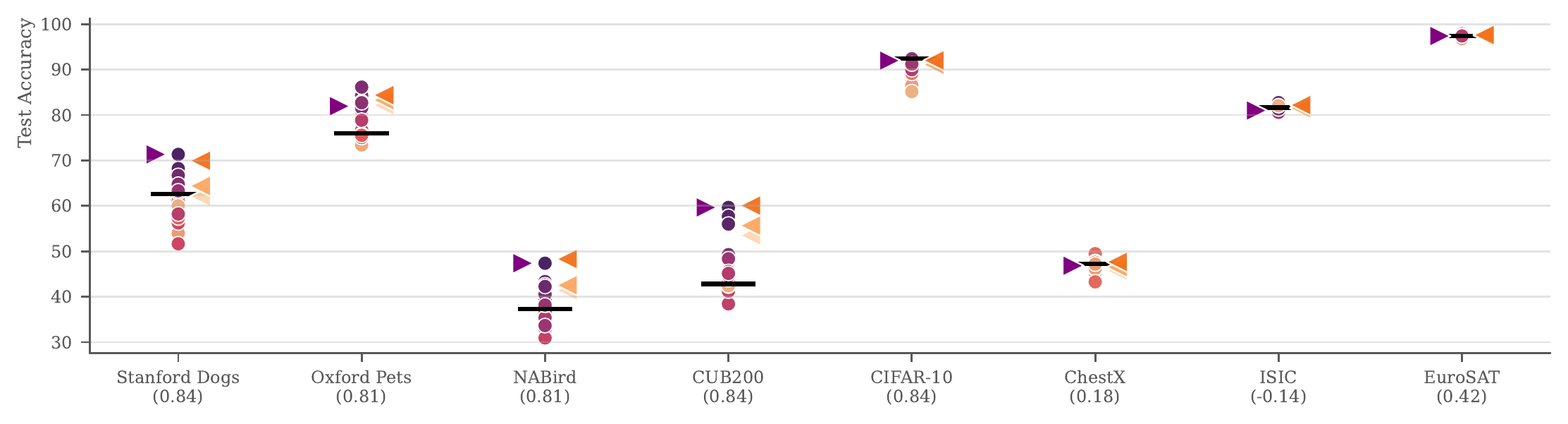}
    \end{adjustbox}
    \caption{Test accuracy for distillation with each dot representing single-source distillation from different source models. The colors represent the task similarity for the source models (from small to large; \gradientcolor). We include the performance from fine-tuning ImageNet (\rule[.15\baselineskip]{1.3em}{1.3pt}), \single{}; i.e. distillation of the highest ranked source model ($\begingroup\color{purple}\blacktriangleright\endgroup$) as well as \multiEqual{} ($\begingroup\color{orange_0}\blacktriangleleft\endgroup$), and \multi{p} where weights are proportional to task similarity with power $p=1$ ($\begingroup\color{orange_1}\blacktriangleleft\endgroup$), and $p=12$ ($\begingroup\color{orange_2}\blacktriangleleft\endgroup$), respectively. The numbers in parentheses at the bottom are Spearman correlations between the task similarity and test accuracy for single-source distillation.}
    \label{fig:natural_images_overall}
\end{figure*}
The aim of this paper is to train an accurate model for a given target task, subject to limited labeled data and computational constraints (\eg limited compute resources).
Formally, we assume that our target task is specified via a small labeled training set $D_{\tau}^{l}$. Furthermore, we assume (a) the availability of a set of unlabeled data, $D_{\tau}^{u}$, associated with the target task, and (b) the ability to perform inference on a set $\S = \{\M_s\}^S_{s=1}$ of $S$ different \emph{source} models, $\M_s$, trained on various source tasks different from the target task, We emphasize that we have no access to any source data   which could be practical due to storage, privacy, and computational constraints. Neither do we need full access to the source models provided we can perform inference on the models anywise (\eg through an API).

We assume that the architecture of the target model, $\M_{\tau}$, must be chosen to meet any applicable computational constraints. This can imply that no suitable target architecture is available in the set of source models $\S$, making classical transfer learning impossible. For simplicity, we restrict our models (regardless of source or target) to classification models that can be parameterized as $\M = h \circ \phi$; the feature extractor $\phi$ embeds input $\bx$ into a feature representation, and the classifier head, $h$, maps the feature $\phi(\bx)$ into predicted conditional class probabilities, $P(\by \mid \bx)$. 

\section{Cross-Task Distillation for Constructing Efficient Models from Foundation Models}
To construct an efficient model, we propose to distill large foundation models. Along this vein, we propose two variants: (a) \single that distills the single nearest source model (Section \ref{sec:single}) and (b) \multi{} that distills a weighted collection of source models (Section \ref{sec:multi}).

\subsection{\single} 
\label{sec:single}
To construct a single efficient target model, \single undergoes two steps sequentially: (a) selecting an appropriate source model and (b) distilling the knowledge from the selected source model into the target model. For ease of exposition, we start by explaining the distillation process and then discuss how to select an appropriate source model. 

\paragraph{Distilling a selected source model.}
Given a selected source model $\M_{s}$, the target model $\M_{\tau} = h_{\tau} \circ \phi_{\tau}$ is trained by minimizing a weighted sum of two loss functions,
\begin{align}\label{eq:single_loss}
    \L_{\mathrm{single}} \eqdef \lambda \L^{\text{labeled}} + (1-\lambda) \L_s^{\text{distill}},
\end{align}
where $\lambda \in [0,1]$. The first loss function is the standard supervised objective over the labeled data,
\begin{align}\label{eq:supervised_loss}
    \L^{\text{labeled}} \eqdef \frac{1}{|\D_\tau^l|}\sum_{(\bx_i, \by_i) \in \D_{\tau}^{l}} \ell_{CE}\left(h_{\tau}(\phi_{\tau}(\bx_i)), \by_i\right),
\end{align}
where $\ell_{CE}(\cdot, \cdot)$ is the cross-entropy loss. The second loss function is a distillation loss over the unlabeled data,
\begin{align}\label{eq:distill_loss}
    \L_s^{\text{distill}} \eqdef \frac{1}{|\D_\tau^u|}\sum_{\bx_i \in \D_{\tau}^{u}} \ell_{CE}\left(h^{s}_{\tau}(\phi_{\tau}(\bx_i)), \M_{s}(\bx_i))\right).
\end{align}
Note, the source and target tasks do not share the same label space so we introduce an additional classifier head, $h^{s}_{\tau}$, which maps the features from the target task feature extractor, $\phi_{\tau}$, to the label space of the source task. This additional classifier head, $h^{s}_{\tau}$, is discarded after training and only the target classifier head, $h_{\tau}$, is used for inference.

In principle, we could add additional semi-supervised losses, such as the FixMatch loss \citep{sohn2020fixmatch} to propagate label information from the labeled set to the unlabeled set for better performance, but this would add additional hyperparameters and entangle the effect of our methods. We leave such explorations to future work.

\paragraph{Selecting the nearest source model for distillation.}
Selecting a source model for distillation is an under-explored problem. 
Given the recent success of using task similarity metrics \citep{Bolya2021ScalableLearning} for selecting foundation models for fine-tuning, we conjecture that high similarities between a source model and the target task could indicate better performance of the distilled model (we verify this in Section \ref{sec: result_single}). However, quantifying similarities between tasks/models is an open research question with various successes \citep{achille2019task2vec, nguyen2020leep}. For simplicity, we pick our similarity based on one simple intuition: target examples with identical labels should have similar source representations and vice versa. Along this vein, the recently introduced metric, PARC \citep{Bolya2021ScalableLearning} fits the bill.

For convenience, we briefly review PARC. Given a small labeled probe set $\D_{\tau}^{p} = \{(\bx_i, \by_i) \}_{i=1}^n \subseteq \D_{\tau}^{l}$ and a source representation of interest $\phi_s$, PARC first constructs two distance matrices $D_{\phi_s}$, $D_Y$ based on the Pearson correlations between every pair of examples in the probe set;
\begin{align*}
    D_{\phi_s} &= 1 - \mathrm{pearson}(\{\phi_s(\bx_i)\}_{i=1}^n),\\
    D_Y &= 1 - \mathrm{pearson}(\{\by_i\}_{i=1}^n).
\end{align*}
PARC is computed as the Spearman correlation between the lower triangles of the distance matrices;
\begin{align*}
    \mathrm{PARC}(\phi_s, Y) = \mathrm{spear}\left(\{D_{\phi_s}[i, j]\}_{i <j}, \{D_{Y}[i, j]\}_{i < j}\right).
\end{align*}
Intuitively, PARC quantifies the similarity of representations by comparing the (dis)similarity structures of examples within different feature spaces: if two representations are similar, then (dis)similar examples in one feature space should stay (dis)similar in the other feature space. In Figure \ref{fig:natural_images_overall} and \ref{fig:expert_metric_correlation} we show that ranking source models by PARC correlates well with test accuracy and that selecting an appropriate source model can yield significant improvements.

\subsection{\multi{}}\label{sec:multi}
Above, \single{} assumes a single optimal source model exists for the target task, but what if no single source model aligns well with our target task? To alleviate this issue, we propose to distill multiple source models, weighted according to their similarities with the target tasks. In the following, we explain our weighted distillation objective and how the weights are constructed. Figure \ref{fig:method_figure} is a schematic depiction of the approach \multi{}.

\begin{table*}[!t]
        \begin{adjustbox}{width=1.2\textwidth,center}
            \centering
\input{tables/vs_finetuned_source_alt_v2.tex}
        \end{adjustbox}
        \caption{Cross-task distillation compared to baselines. MobileNetV3 models (target architecture) trained with our methods are highly competitive with baseline methods on MobileNetV3 as well as baseline methods for more demanding model architectures (source architectures: Alexnet, GoogLeNet, ResNet-18, ResNet-50). We highlight the top 3 methods, which comply with compute requirements (i.e. MobileNetV3) for each target task by \textbf{bold}, \underline{underline}, and \textit{italic}, respectively. We also indicate the target data used by different methods.}
        \label{tbl:vs_finetuned_source_models}
\end{table*}

\paragraph{Weighted objective for distilling multiple sources. }
Given a set of source models $\S = \{M_{s}\}_{s=1}^S$, we modify the distillation loss of \eqref{eq:single_loss} with a weighted sum of multiple distillation losses (one for each source model):
\begin{align}
    \label{eq:multi_source_objective}
    \L_{\mathrm{multi}} \eqdef \lambda \L^{\text{labeled}} + (1-\lambda) \sum_{s=1}^S \alpha_s \L_s^{\text{distill}},
\end{align}
where $\lambda, \alpha_1, \dots, \alpha_S \in [0,1]$ ($ \L^{\text{labeled}}$ and $\L_s^{\text{distill}}$ are as defined in \eqref{eq:supervised_loss} and \eqref{eq:distill_loss}, respectively). Here $\alpha_s$ is the relative weight assigned to each source model such that $\sum_{s=1}^S \alpha_s = 1$. Once again, we could add additional semi-supervised losses, such as the FixMatch loss, but to ensure simplicity, we leave such explorations for future research.

\paragraph{Task similarity weighting of source models}
Simply assigning equal weight to all source models is sub-optimal (e.g. weighing source models trained on ImageNet and Chest X-ray equally might not be optimal for recognizing birds). As such, we propose to compute the source weight $\alpha_s$ from a task similarity metric between the $s$-th source model and the target task. In particular, let $e_s$ be such a similarity metric, then we compute the source weights $\{\alpha_i\}_{i \in [S]}$ as 
\begin{align}
    \label{eq:metric_weight}
    \alpha_i = \frac{\underline{e}_i^{p}}{\sum_{s=1}^{S} \underline{e}_s^{p}}, \quad \text{where } \underline{e}_j = \mathrm{max}(0, e_j)
\end{align}
for $j = 1, \dots, S$. Here $p$ is a hyperparameter to re-scale the distribution of the weights. Larger $p$ assigns more weight to the most similar source models, while $p=0$ corresponds to equal weights for all models (denoted \multiEqual{}), and $p \to \infty$ assigns all weight to the most similar source model (i.e. \single{}). When relevant, we use the notation \multi{p} to indicate the choice of $p$.

\paragraph{Scalability}
For \multi{} to be feasible, compared to \single{}, we need to ensure that the training procedure scales well with the size of $\S$. Since the computation of the weights $\{\alpha_s\}_{s=1}^{S}$ is based on the small probe set and is almost identical to the selection procedure for \single{} this is a negligible step. When training the target model, we merely require one forward pass on the unlabeled target dataset with each source model (to obtain pseudo-labels) as well as training of a one-layer classifier head per source model, both of which are cheap compared to the full training procedure of the target model. Nonetheless, one could employ a pre-selection of the top-$k$ source models with the largest task similarity, thereby reducing the number of classifier heads and forward passes required. However, doing so introduces another hyperparameter, $k$, (i.e. how many models to use) complicating the analysis. Moreover, since large $p$ induces such pre-selection in a \textit{soft} manner,
we leave it to future research to determine how to select the appropriate $k$.

%% file: tables/vs_finetuned_source_alt_v2.tex
\fontsize{7pt}{7pt}\selectfont
\newcolumntype{C}{>{\centering\arraybackslash}X}
\newcolumntype{P}{>{\centering\arraybackslash}}
\setlength{\tabcolsep}{0pt}
\setlength{\extrarowheight}{5pt}
\renewcommand{\arraystretch}{0.75}
\begin{tabularx}{1.2\linewidth}{p{1.5cm}!{\color{lightgray}\vline\hspace{0.2cm}}p{3.2cm}!{\color{lightgray}\vline\hspace{0.2cm}}CC!{\color{lightgray}\vline\hspace{0.2cm}}CCCCCCCC !{\color{lightgray}\vline} C}
\toprule
 & & \multicolumn{2}{C!{\color{lightgray}\vline\hspace{0.2cm}}}{\rotatebox{0}{\shortstack[c]{Target Data\\ Labeled\;\; Unlabeled}}} & \rotatebox{90}{CIFAR-10} & \rotatebox{90}{CUB200} & \rotatebox{90}{ChestX} & \rotatebox{90}{EuroSAT} & \rotatebox{90}{ISIC} & \rotatebox{90}{NABird} & \rotatebox{90}{\shortstack[c]{Oxford\\Pets}} & \rotatebox{90}{\shortstack[c]{Stanford\\Dogs}} & \rotatebox{90}{Mean} \\
\midrule
\multirow[c]{6}{*}{\shortstack[l]{MobileNetV3\\ \fontsize{5.5pt}{5.5pt}\selectfont(0.24 GFLOPs)}} & IN+Transfer & \checkmark & - & \underline{92.4} & 42.8 & \underline{47.3} & 97.4 & 81.6 & 37.3 & 75.9 & 62.6 & 67.2 \\
 & IN+FixMatch & \checkmark & \checkmark & \textbf{93.5} & 41.9 & 38.5 & \textbf{98.1} & \textbf{82.6} & \textit{42.8} & \underline{83.4} & \textit{65.8} & 68.3 \\
 \cline{2-13}
 & \randomSingle & \checkmark & \checkmark & 89.6 & 46.5 & 46.6 & 97.4 & \textit{81.8} & 39.0 & 79.4 & 61.9 & 67.8 \\
& \textbf{(Ours)} \single & \checkmark & \checkmark & 92.0 & \underline{59.6} & 46.8 & 97.4 & 81.0 & \underline{47.4} & 81.9 & \textbf{71.3} & \underline{72.2} \\
 \cline{2-13} 
 & \multiEqual{} & \checkmark & \checkmark & 90.8 & \textit{53.5} & 45.7 & 97.5 & 81.5 & 41.4 & \textit{82.1} & 62.1 & \textit{69.3} \\
& \multiRandom{} & \checkmark & \checkmark & 87.9 & 44.9 & \textit{46.9} & \underline{97.8} & 81.6 & 39.6 & 80.2 & 59.2 & 67.3 \\
 & \textbf{(Ours)} \multi{} & \checkmark & \checkmark & \textit{92.0} & \textbf{60.0} & \textbf{47.7} & \textit{97.6} & \underline{82.2} & \textbf{48.3} & \textbf{84.4} & \underline{69.9} & \textbf{72.8} \\
 \midrule
 \midrule
\multirow[c]{2}{*}{\shortstack[l]{AlexNet\\ \fontsize{5.5pt}{5.5pt}\selectfont(0.71 GFLOPs)}} & IN+Transfer & \checkmark & - & 85.0 & 18.4 & 46.2 & 91.9 & 67.8 & 13.0 & 50.9 & 29.1 & 50.3 \\
 & Fine-tune Selected Source & \checkmark & - & 88.0 & 30.4 & 42.9 & 89.8 & 74.5 & 17.9 & 66.8 & 41.3 & 56.5 \\
\midrule
\multirow[c]{2}{*}{\shortstack[l]{GoogLeNet\\ \fontsize{5.5pt}{5.5pt}\selectfont(1.51 GFLOPs)}} & IN+Transfer & \checkmark & - & 91.8 & 42.8 & 41.4 & 96.8 & 80.5 & 36.5 & 84.8 & 65.9 & 67.6 \\
 & Fine-tune Selected Source & \checkmark & - & 91.6 & 61.2 & 48.6 & 96.9 & 78.3 & 33.0 & 87.8 & 71.8 & 71.2 \\
\midrule
\multirow[c]{2}{*}{\shortstack[l]{ResNet-18\\ \fontsize{5.5pt}{5.5pt}\selectfont(1.83 GFLOPs)}} & IN+Transfer  & \checkmark & - & 92.2 & 37.8 & 45.2 & 96.6 & 80.2 & 34.0 & 80.2 & 58.2 & 65.6 \\
 & Fine-tune Selected Source & \checkmark & - & 91.3 & 58.2 & 46.4 & 97.0 & 75.8 & 35.4 & 80.7 & 69.3 & 69.3 \\
\midrule
\multirow[c]{2}{*}{\shortstack[l]{ResNet-50\\ \fontsize{5.5pt}{5.5pt}\selectfont(4.14 GFLOPs)}} & IN+Transfer &\checkmark & - & 92.9 & 42.0 & 43.4 & 96.8 & 79.9 & 39.9 & 83.3 & 65.9 & 68.0 \\
 & Fine-tune Selected Source & \checkmark & - &93.0 & 70.8 & 43.9 & 97.2 & 81.3 & 47.4 & 84.8 & 79.3 & 74.7 \\
\bottomrule
\end{tabularx}

%% file: 4_experiments.tex
\section{Experiments and Results}\label{sec:experiments}
\subsection{Experimental Setup}
\textbf{Benchmark.}
Despite our methods being designed with the interest of using large vision models (that are potentially only available for inference), such a setting is intractable for our research. Thus, to allow for controlled experimentation we restrict our source models to a more tractable scale.
In particular, we modify an existing transfer learning benchmark: Scalable Diverse Model Selection by \cite{Bolya2021ScalableLearning}, and use the publicly available models to construct a set of source models for each target task. Thus, we consider a set consisting of 28 models: 4 architectures (AlexNet, GoogLeNet, ResNet-18, and ResNet-50 \citep{krizhecsky2012imagenet, He2016DeepRecognition}) trained on 7 different source tasks (CIFAR-10, Caltech101, CUB200, NABird, Oxford Pets, Stanford Dogs, and VOC2007). For the target tasks, we consider 8 different tasks covering various image domains (Natural images: CIFAR-10, CUB200, NABird, Oxford Pets, Stanford Dogs; X-ray: ChestX; Skin Lesion Images: ISIC; Satellite Images: EuroSAT). We carefully remove any source models associated with a particular target task, if such exists, in order to avoid information leakage between source and target tasks (see also supplementary materials for further considerations). 
For the target architecture, we use MobileNetV3 \citep{Howard2019SearchingMobileNetV3} due to its low computational requirements compared to any of the source models. We refer the reader to the supplementary material for further details on implementation.

\textbf{Baselines.} 
We consider a set of different baselines: based on ImageNet initializations we consider \transfer{} (fine-tunes ImageNet representations using only the labeled data), and \fixmatch{} \citep{sohn2020fixmatch} (fine-tunes the ImageNet representation using labeled and unlabeled data), and based on source model initializations we fine-tune the highest-ranked source model of each source architecture. To show the importance of using the right source model(s) to distill, we also compare \single{} to \randomSingle{} which is the average of distilling from a randomly selected source, and for comparison to \multi{} we also construct distilled models using the multi-source objective \eqref{eq:multi_source_objective} with a random weight (\multiRandom{}) and equal weights (\multiEqual{}). For ease of exposition, we present results for \single{} (Section \ref{sec: result_single}) and \multi{} (Section \ref{sec: result_multi}) in separate sections.

\subsection{Results for \single{}}
\label{sec: result_single}
We compare \single{} with the baselines in Table \ref{tbl:vs_finetuned_source_models} and Figure \ref{fig:natural_images_overall}. 
Our observations are as follows.

\textbf{Distillation with the right source model is better than fine-tuning from ImageNet.} We observe that within the same target architecture (MobileNetv3), simply fine-tuning ImageNet representations (\transfer{}) is less optimal than distilling from the most similar single model (\single{}). In fact, for fine-grained datasets such as CUB200, NABird, Oxford Pets, and Stanford Dogs, we observe that distilling from an appropriate source model (\single{}) could yield much better performance than fine-tuning from a generalist ImageNet representation. More surprisingly, even with the aid of unlabeled data, models fine-tuned from ImageNet representations using a label propagation style approach (\fixmatch{}) still underperform distillation-based methods by at least 3.9\% on average. These observations indicate the importance of selecting the right source model for transfer/distillation. 

\textbf{Distilling to efficient architecture could be better than fine-tuning larger models.} In Table \ref{tbl:vs_finetuned_source_models}, we include the performance when fine-tuning larger architectures trained on ImageNet (\transfer{}) and the source model (of the same architecture) most similar to each target task selected using PARC (\textsc{Fine-tune Selected Source}). A few observations are immediate: (a) our choice of task similarity metric is effective for transfer; across all 4 architectures, we observe at least 4\% improvement over simple fine-tuning from ImageNet, which validates the results by \citet{Bolya2021ScalableLearning}, and (b) with the aid of unlabeled data and distillation, the computationally efficient architecture MobileNetV3 can outperform larger architectures fine-tuned on labeled data from the target task (i.e. AlexNet, GoogLeNet, ResNet-18). Although underperforming fine-tuning a ResNet-50 initialized with the most similar ResNet-50 source model by a mere average of 2.5\%-points (\textsc{Fine-tune Selected Source}), using a ResNet-50 would require $17.5\times$ more computations during inference to achieve such improvements.

\subsubsection{Task Similarity Metrics for \single{}}
One key component of \single{} is to select the source model to perform cross-task distillation on using task similarity metrics. Despite many many existing metrics for quantifying task similarities, their effectiveness for distillation remains unclear. Given the myriads of metrics, we restrict our focus to metrics that can capture similarities between a source representation of a target example and its one-hot label representation. Along this vein, two questions arise: which metric to use for comparing representations, and which representations from a source model should be used to represent a target example?

For the first question, we look into multiple metrics in the literature that compares various representations: CKA \citep{Cortes2012AlgorithmsAlignment}, RSA \citep{Dwivedi2019RepresentationLearning}, and PARC \citep{Bolya2021ScalableLearning}. For the second question, we look into the common representations from a source model: the features $\phi$ and the probabilistic outputs $h \circ \phi$.

To establish the effectiveness of our choice of similarity metric, we report the Spearman correlation between the task similarities and the test accuracy of the distilled models in Table \ref{tbl:single_teacher_spearman_corr}. We see that features from the source models can better capture the correlation between the source models and the test accuracy of the distilled models, than the probabilistic pseudo-labels. In addition, we also see a much higher correlation among natural tasks (compared to specialized tasks such as ChestX, EuroSAT, and ISIC) which suggests that our choice of task similarity is effective at selecting similar tasks. Besides, we also observe a higher correlation when using PARC compared to the other metrics, thus validating our choice of using PARC as the default metric. 

\begin{table*}[t]
        \centering
        \input{tables/single_teacher_spearman_corr}

        \caption{Spearman correlation between test accuracy after all possible single-source distillations and task similarities associated with the source models.
        Generally feature representations correlate better with distillation performance compared to pseudo-label representations.}
        \label{tbl:single_teacher_spearman_corr}
\end{table*}

To further establish the effectiveness of our metrics to rank various source models, we compute the relative test accuracy between the top-3 models most similar to the target task and the top-3 best models after distillation (see Table \ref{tbl:single_teacher_relative_acc_top_3}). Again, we observe that all three metrics are capable of ranking affinity between source models, but ranking the models with PARC outperforms the other two metrics. 

\begin{table*}[t]
        \centering
        \input{tables/single_teacher_relative_acc_top_3}

        \caption{Relative accuracy of top-3 single-source distilled models selected by task similarity over the average of the 3 actual best models. We compute the average test accuracy of the top-3 highest ranked target models and divide it by the average of the 3 actually best-performing target models.}
        \label{tbl:single_teacher_relative_acc_top_3}
\end{table*}

\begin{figure*}[htbp]
    \begin{minipage}{0.48\textwidth}
        \centering
        \includegraphics[width=\linewidth]{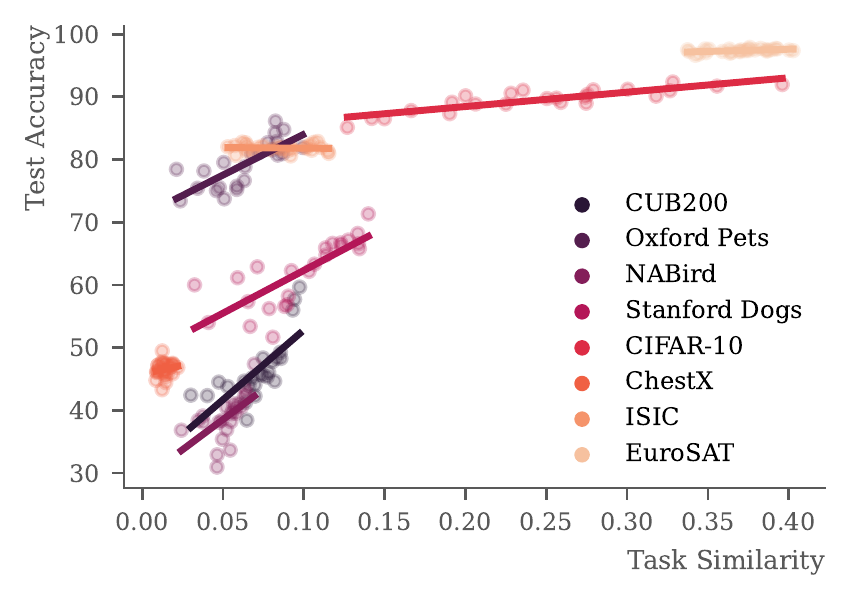}
        \caption{Test accuracy of single-source distillation and raw task similarity score using PARC on the feature representations. The scores are on different scales for different tasks, but almost all tasks have a positive correlation between test accuracy and task similarity.}
        \label{fig:expert_metric_correlation}
    \end{minipage}
    \hfill
    \begin{minipage}{0.48\textwidth}
        \centering
        \includegraphics[width=\linewidth]{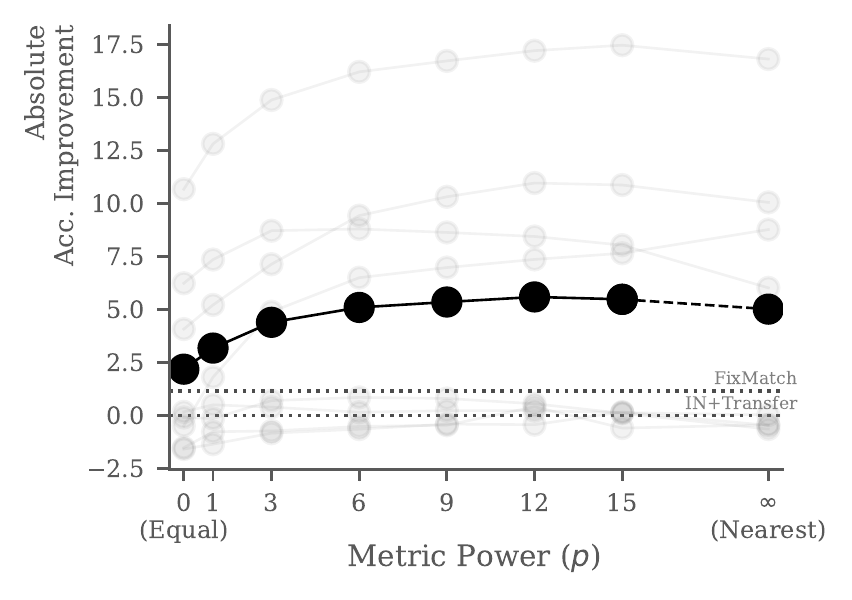}
        \caption{Improvement over \transfer{}. Here $\bullet$ is the average improvement over all eight target tasks and $\begingroup\color{black!20}\bullet\endgroup$ represents the performance on a target task. Note, $p = 0$ corresponds to \multiEqual{}, and $p=\infty$ corresponds to \single{}.}
        \label{fig:abs_improvement_over_transfer}
    \end{minipage}
\end{figure*}

\subsection{Results for \multi{}}
\label{sec: result_multi}

From Table \ref{tbl:vs_finetuned_source_models}, we observe that \multi{} compares favorably to \single{}, thus the conclusions for \single{} translates to \multi{}. Yet, one particular task, Oxford Pets, is worth more attention.  On Oxford Pets (classification of different breeds of cats and dogs), we observe that distilling from multiple weighted sources (\multi{}) is much better than distilling from the single most similar source (\single{}), which is a ResNet-18 trained on Caltech101 (that can recognize concepts such as Dalmatian dog, spotted cats, etc.). Although the most similar source model contains relevant information for recognizing different breeds of dogs and cats, it might not contain all relevant knowledge from the set of source models that could be conducive to recognizing all visual concepts in Oxford Pets. In fact, we observe that the second most similar model is a GoogLeNet model trained on Stanford Dogs to recognize more dog breeds than the most similar source model (but incapable of recognizing cats). In this case, \multi{} allows aggregation of knowledge from multiple sources and can effectively combine knowledge from different source models for a more accurate target model than distillation from a single source. This suggests that \emph{under certain conditions such as high heterogeneity in data, distilling from multiple source models can outperform distilling a single best source model.}

\subsubsection{Task Similarity Metrics for Weighing Sources}
We have established that our task similarity metric can capture the correlation between the source model representations and the test accuracy of the distilled models. However, it is not a priori clear that weighing source models based on the ranking of their affinity to the target task would yield better performance for multi-source distillation. As such, we investigate alternative choices of weighing schemes for a subset of 5 target tasks (CUB200, EuroSAT, ISIC, Oxford Pets, Stanford Dogs): \textsc{Inverse} (weights are inversely proportional to task similarity), \multiRandom{} (weights are sampled uniformly on a 4-simplex), \randomSingle (randomly selecting a single source model), and \multiEqual{} (equal weights for all models).

Through Figure \ref{fig:alt_source_weights}, we find that distilling from a single or set of source models ranked using the similarity metric is much more effective than distilling from source models that are weighted randomly or equally (\multiRandom{} or \multiEqual{}). In addition, the fact that \textsc{Inverse} underperforms \transfer{} on average suggests that it is crucial to follow the ranking induced by the similarity metrics when distilling the sources and that the metric ranks both the most similar source models and the least similar source models appropriately.

\subsubsection{Effect of $p$}
Our task similarity metrics give a good ranking of which source models to select for distillation but it is unclear whether the similarity score could be used directly without any post-processing. To investigate, we visualize the relationship between the test accuracy of the models distilled from a single source and our task similarity. From Figure \ref{fig:expert_metric_correlation}, it is clear that the distribution of task similarities depends on the target task, which motivates our normalization scheme.

In addition, it is not apriori clear that the weights should scale linearly with the similarity scores. Thus, we investigate the effect of the rescaling factor, $p$,  for constructing the weights. In Figure \ref{fig:abs_improvement_over_transfer}, we see that although no rescaling ($p=1$) outperforms equal weighting, it is less optimal than \eg $p=12$ (our default). This suggests that task similarity and good weights have a monotonic, but non-linear relationship.

\subsection{Additional Ablations and Analyses}
Due to space constraints, we 
include additional ablations and analyses in the supplementary materials. We summarize the main findings as follows.

\mypara{ResNet-50 as target model.}
Averaged over 8 tasks, \multi{} outperforms both \transfer{} and \multiEqual{} by 5.6\% and 3.8\%, respectively. Also, compared to ImageNet initialization, using \multi{} with the most similar ResNet-50 source model as target model initialization improves accuracy by 1.0\%.

\mypara{Improvements on VTAB.}
\multi{} outperforms \transfer{} averaged over the \vtabNatural and \vtabSpecialized tasks of VTAB, by 5.1\% and 0.8\%, respectively. \single{} outperform by 4.8\% and 0.6\%, respectively.

\mypara{Fewer labels.}
\multi{} and \single{} outperform \transfer{} (by 6.8\% and 4.4\%, respectively) under a setup with even fewer labeled samples.

\mypara{Additional analysis of task similarity metrics.}
We consider additional correlation metrics and top-$k$ relative accuracies of the selected models --- all supporting the usefulness of task similarity to weigh and select source models.

%% file: tables/single_teacher_spearman_corr.tex
\fontsize{7pt}{7pt}\selectfont
\newcolumntype{C}{>{\centering\arraybackslash}X}
\setlength{\tabcolsep}{0pt}
\setlength{\extrarowheight}{7pt}
\renewcommand{\arraystretch}{0.75}
\begin{tabularx}{1\linewidth}{p{0.4cm}p{0.9cm}!{\color{lightgray}\vline} CCCCCCCC !{\color{lightgray}\vline} C}
\toprule
 &  & \rotatebox{90}{CIFAR-10} & \rotatebox{90}{CUB200} & \rotatebox{90}{ChestX} & \rotatebox{90}{EuroSAT} & \rotatebox{90}{ISIC} & \rotatebox{90}{NABird} & \rotatebox{90}{Oxford Pets} & \rotatebox{90}{Stanford Dogs} & \rotatebox{90}{Mean} \\
\midrule
\multirow[c]{3}{*}{\rotatebox{90}{Pseudo}} & CKA & 0.72 & 0.62 & 0.23 & 0.39 & -0.04 & 0.31 & 0.69 & 0.11 & 0.38 \\
 & PARC & 0.79 & 0.79 & 0.02 & 0.17 & 0.06 & 0.48 & 0.72 & 0.54 & 0.45 \\
 & RSA & 0.82 & 0.31 & -0.11 & 0.30 & \textbf{0.10} & -0.03 & 0.65 & 0.38 & 0.30 \\
\midrule
\multirow[c]{3}{*}{\rotatebox{90}{Feature}} & CKA & 0.82 & 0.39 & \textbf{0.36} & 0.21 & -0.04 & 0.47 & 0.69 & 0.55 & 0.43 \\
 & PARC & 0.84 & \textbf{0.84} & 0.18 & \textbf{0.42} & -0.14 & \textbf{0.81} & 0.81 & 0.84 & \textbf{0.58} \\
 & RSA & \textbf{0.86} & 0.81 & 0.03 & 0.38 & 0.03 & 0.28 & \textbf{0.89} & \textbf{0.85} & 0.52 \\
\bottomrule
\end{tabularx}

%% file: tables/single_teacher_relative_acc_top_3.tex
\fontsize{7pt}{7pt}\selectfont
\newcolumntype{C}{>{\centering\arraybackslash}X}
\setlength{\tabcolsep}{0pt}
\setlength{\extrarowheight}{7pt}
\renewcommand{\arraystretch}{0.75}
\begin{tabularx}{1\linewidth}{p{0.4cm}p{0.9cm}!{\color{lightgray}\vline} CCCCCCCC !{\color{lightgray}\vline} C}
\toprule
 &  & \rotatebox{90}{CIFAR-10} & \rotatebox{90}{CUB200} & \rotatebox{90}{ChestX} & \rotatebox{90}{EuroSAT} & \rotatebox{90}{ISIC} & \rotatebox{90}{NABird} & \rotatebox{90}{Oxford Pets} & \rotatebox{90}{Stanford Dogs} & \rotatebox{90}{Mean} \\
\midrule
\multirow[c]{3}{*}{\rotatebox{90}{Pseudo}} & CKA & 99.1 & 95.6 & 97.4 & 99.6 & 98.8 & 89.4 & \textbf{100.0} & 97.6 & 97.2 \\
 & PARC & 99.5 & \textbf{100.0} & 95.5 & 99.6 & 98.5 & 99.7 & 98.8 & \textbf{99.7} & 98.9 \\
 & RSA & \textbf{100.0} & 77.7 & 96.5 & 99.7 & 98.5 & 87.2 & 98.6 & 97.6 & 94.5 \\
\midrule
\multirow[c]{3}{*}{\rotatebox{90}{Feature}} & CKA & \textbf{100.0} & 95.6 & 97.0 & \textbf{99.8} & \textbf{99.0} & 93.3 & \textbf{100.0} & 96.4 & 97.6 \\
 & PARC & \textbf{100.0} & \textbf{100.0} & \textbf{97.8} & 99.7 & 98.3 & \textbf{100.0} & 97.1 & 98.5 & \textbf{98.9} \\
 & RSA & \textbf{100.0} & \textbf{100.0} & 96.7 & 99.8 & 98.9 & 94.9 & 98.9 & 98.8 & 98.5 \\
\bottomrule
\end{tabularx}

%% file: 5_conclusion.tex
\section{Conclusion}
We investigate the use of diverse source models to obtain efficient and accurate models for visual recognition with limited labeled data.
In particular, we propose to distill multiple diverse source models from different domains weighted by their relevance to the target task without access to any source data.
We show that under computational constraints and averaged over a diverse set of target tasks, our methods outperform both transfer learning from ImageNet initializations and state-of-the-art semi-supervised techniques.

%% file: 6_supplementary_materials.tex
\appendix
\clearpage

\section{Additional Ablations and Analyses}\label{app:additional_results}
We present additional results and implementation details in the supplementary. To avoid confusion, we use the same set of index numbers as in the main text to refer to the tables and figures. Please find Tables 1-3 and Figures 1-6 in the main text.

\subsection{Results on VTAB}
We report the results of our VTAB \citep{Zhai2019ABenchmark} experiment in Table \ref{tbl:vtab_full}. On VTAB, We find that both \multi{} and \single{} distillation outperform \transfer{} on each of the \vtabNatural tasks. Particularly, \multi{} outperforms \transfer{} with $13.9\%$-points on CIFAR-10 and $10.6\%$-points on Sun397 and averaged across \vtabNatural \multi{} outperforms \transfer{} with $5.1\%$-points. Average over \vtabSpecialized both \multi{} and \single{} outperform \transfer{}, although with a small margin. Finally, averaged over \vtabStructured \transfer{} outperforms our methods, but due to the nature of these tasks, we do not expect source models to transfer well to these tasks.\footnote{The \vtabStructured tasks are mainly (ordinal) regression tasks transformed into classification tasks, and thus it seems reasonable to expect very general features (such as those from an ImageNet pre-trained model) to generalize better to such constructed tasks than specialized source models.} Yet, we still obtain the best accuracy on DMLab, dSpr-Loc, and sNORB-Azimuth.

\begin{table*}[htbp]
    \centering
    \input{tables/vtab/full}
    \caption{Top-1 accuracy by dataset in VTAB. The accuracy for each task is in grey, and the average accuracy for each category of tasks is in black. Note, the \protect\vtabMean is the average across all tasks, not categories. The largest value in each column is marked in bold. Here \multi{} is with $p=9$.}
    \label{tbl:vtab_full}
\end{table*}

\subsection{Relative accuracy of single-source distillation}
Similarly to Table \ref{tbl:single_teacher_relative_acc_top_3}, we extend our evaluation of how well the task similarity selects the best source models for single-source distillation. We report the ratio between the average test accuracy of the top-$k$ target models ranked using the task similarity and the average test accuracy for the actual top-$k$ target models found after the fact in Table \ref{tbl:single_teacher_relative_acc_top_1}, Table \ref{tbl:single_teacher_relative_acc_top_3_alt}, and Table \ref{tbl:single_teacher_relative_acc_top_5} for $k=1$, $k=3$, and $k=5$, respectively.

We find that generally, using task similarity on feature representations rather than the corresponding pseudo-labels yields better rankings, but also that PARC shows very little difference between features and pseudo-labels for all considered $k \in \{1, 3, 5\}$.

\begin{figure}[!t]
    \centering
    \includegraphics[width=0.5\linewidth]{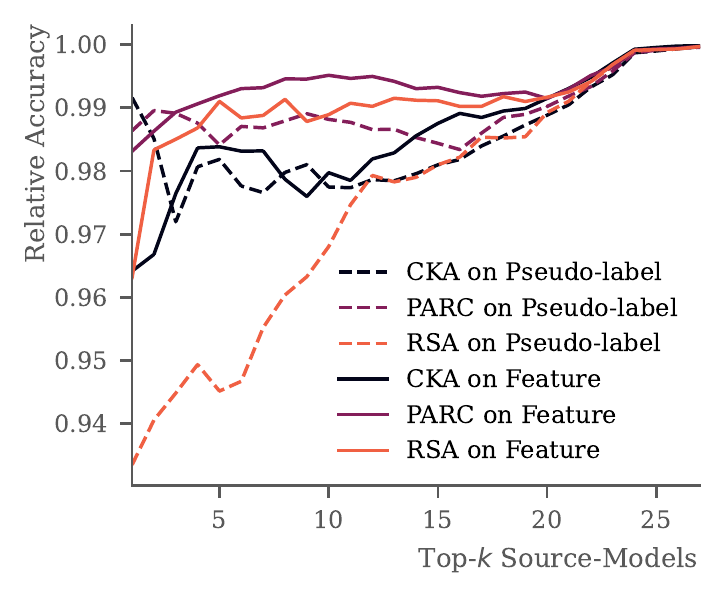}
    \captionof{figure}{Relative accuracy of top-$k$ single-source distilled target models selected by task similarity over the average of the top-$k$ actual best target models found in hindsight. If the ordering by task similarity were perfectly correct, the relative accuracy would be $1$ for all $k$. See Table \ref{tbl:expert_metric_all_relative_acc} for the average of each metric across all $k$.}
    \label{fig:expert_metric_relative_acc}
\end{figure}

\paragraph{Relative accuracy over all $k$.} The relative accuracy measure reported above is sensitive to $k$ and the actual accuracy values of the models. I.e. if a metric flips the order of the best and second best model when there is a notable performance gap between the two models, the relative accuracy for $k = 1$ will be low, and we might be mistaken to believe the metric is not working well. However, the metric might rank every model for $k>2$ perfectly correct, and since we typically utilize the full set of source models, the initial mistake should not be detrimental to the selection of the task similarity metric. Thus, in Figure \ref{fig:expert_metric_relative_acc} we plot the relative accuracy for each task similarity metric and all $k \in \{1, \dots, S\}$. We find that while PARC on feature representations is outperformed by both PARC and CKA on pseudo-labels for $k < 3$, PARC on feature representations outperforms all the other metrics for $k \geq 3$. In particular, from Table \ref{tbl:expert_metric_all_relative_acc} we have that on average over all $k < S$, PARC, performs the best.

\begin{table*}[htbp]
    \centering
    \input{tables/single_teacher_relative_acc_top_1}

    \caption{Relative accuracy of top-1 single-source distilled target model selected by task similarity over the best model found in hindsight. We compute the test accuracy of the highest-ranked target model (ranked by some task similarity) and divide this by the test accuracy of the best-performing target model.}
    \label{tbl:single_teacher_relative_acc_top_1}
\end{table*}

\begin{table*}[htbp]
    \centering
    \input{tables/single_teacher_relative_acc_top_3}
    \caption{(Identical to Table \ref{tbl:single_teacher_relative_acc_top_3}) Relative accuracy of top-3 single-source distilled target models selected by task similarity over the average of the 3 best models found in hindsight. We compute the average test accuracy of the top-3 highest ranked target models and divide this average by the average test accuracy of the 3 best-performing target models.}
    \label{tbl:single_teacher_relative_acc_top_3_alt}
\end{table*}

\begin{table*}[htbp]
    \centering
    \input{tables/single_teacher_relative_acc_top_5}

    \caption{Relative accuracy of top-5 single-source distilled target models selected by task similarity over the average of the 5 best models found in hindsight. We compute the results analogously to Table \ref{tbl:single_teacher_relative_acc_top_3_alt} with $k=5$.}
    \label{tbl:single_teacher_relative_acc_top_5}
\end{table*}

\begin{table*}[htbp]
    \centering
    \input{tables/expert_metric_all_relative_acc.tex}

    \captionof{table}{The mean relative accuracy, across all $k$, for each metric in Figure \ref{fig:expert_metric_relative_acc}. The average is bounded in $(0,1]$, and $1$ corresponds to perfect ordering by task similarity. We find that using feature representations consistently outperforms pseudo-labels and that for both feature representations and pseudo-labels PARC performs the best.}
    \label{tbl:expert_metric_all_relative_acc}
\end{table*}

\subsection{Ablation of \textit{p} for \multi{}}
We report the values associated with Figure \ref{fig:abs_improvement_over_transfer} for each target task and all considered choices of $p$ in Table \ref{tbl:multi_source_over_baseline}.
\begin{table*}[t]
    \centering
    \input{tables/general_results.tex}

    \caption{Test accuracy of \multi{} with various choices of $p$, compared to the baseline methods of \transfer{} and \fixmatch{}. We highlight the largest value for each target task in \textbf{bold}, and the results are also visualized in Figure \ref{fig:abs_improvement_over_transfer}.}
    \label{tbl:multi_source_over_baseline}
\end{table*}

\subsection{\multi{} with ResNet-50 as target architecture}
In the main part of the article, we consider the computationally constrained setting, where some compute budget restricts the possible size of our target model. Thus, we use MobileNetV3 models as target models throughout the main paper. However, in Table \ref{tbl:resnet50_multi_source_over_finetune} we remove the computational budget and allow the target model to be of any architecture, and particularly we use a ResNet-50 as the target model. 

We compare \multi{} (with $p=0$ and $p=12$) initialized with either ImageNet pre-trained weights or the weights of the highest ranked ResNet-50 source model to \transfer and \textsc{Fine-tune Selected Source}. We find that \multi{} initialized from ImageNet outperforms \transfer on average for both equal weighting and $p=12$, but underperforms \textsc{Fine-tune Selected Source} for both $p$. However, since \textsc{Fine-tune Selected Source} is initialized from well-selected source model weights, the comparison is not entirely fair. Thus, we also consider the case where we initialize the target model for \multi{} with the weights of the highest ranked ResNet-50 source model, and find that for $p=12$ \multi{} performs on par with \textsc{Fine-tune Selected Source}.

\begin{table*}[t]
    \centering
    \input{tables/resnet50_vs_finetuned_source.tex}
    \caption{\multi{} with ResNet-50 as target model architecture. We compare fine-tuning of the highest ranked source model \citep{Bolya2021ScalableLearning} with \multi{} to both ImageNet-initialized target models and target models initialized from the highest ranked ResNet-50 source model. For $p=12$, \multi{} performs on par with fine-tuning the selected source model. The largest value for each target task is in \textbf{bold}.}
    \label{tbl:resnet50_multi_source_over_finetune}
\end{table*}

\subsection{Normalization of task similarity for source model weighting}
\begin{figure*}[t]
    \centering
    \includegraphics[width=\linewidth]{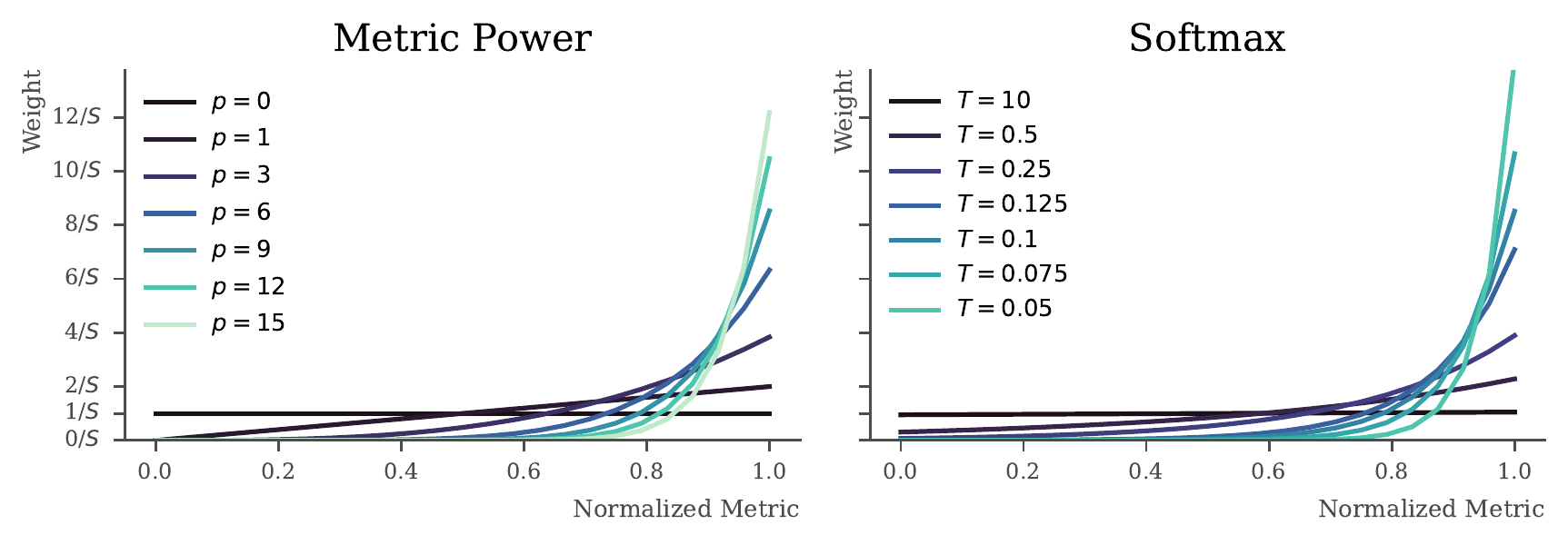}
    \caption{Transformation of weights for various choices of power (left) or softmax temperature (right). Here $S$ is the number of source models, and we consider equidistantly distributed normalized metrics.}
    \label{fig:metric_transformation}
\end{figure*}

We propose to choose the weights $\bm{\alpha} = (\alpha_1, \dots, \alpha_S)$ as
\begin{align*}
    \alpha_i = \frac{\underline{e}_i^{p}}{\sum_{s=1}^{S} \underline{e}_s^{p}}, \quad \text{where} \qquad \underline{e}_j = \mathds{1}_{(e_j > 0)} \;e_j 
\end{align*}
for $j = 1, \dots, S$, and $e_s$ is the task similarity for source model $\M_s$, evaluated on the target task, normalized to satisfy $e_s \in [0,1]$ with min-max normalization over all $e_s$. Here, the hyperparameter, $p$ can be used to increase/decrease the relative weight on the highest ranked source models, with the extremes $p=0$ and $p \to \infty$ corresponding to equal weight and single-source distillation, respectively.
An alternative way to obtain our normalization is to use the softmax function on the task similarities,
\begin{align*}
    \alpha_i = \frac{\exp{\left(\frac{e_i}{T}\right)}}{\sum_{s=1}^{S} \exp{\left(\frac{e_s}{T}\right)}}.
\end{align*}
This does not require clipping the task similarity at $0$, and with the temperature, $T$, we can adjust the relative weight on particular source models. Here, large $T$ flattens the weights, and $T \to \infty$ corresponds to an equal weighting of all source models, while small $T$ increases the weight on the highest-ranked source models. Quantitatively, the two normalization methods can yield similar transformations with appropriate choices of $p$ and $T$ - see Figure \ref{fig:metric_transformation}.

\subsection{Smaller amount of labeled data}
We now repeat the experiment of the main paper across the 8 target datasets with a reduced amount of labeled samples. Here, we reduce the number of labeled samples to $5\%$ (rather than $20\%$) of the training set and report the accuracy in Table \ref{tbl:5_labeled_results}. We find a similar pattern as observed in the main experiment, where \multi{} distillation on average outperforms \transfer irrespective of the choice of $p$. For $p=9$ \multi{} outperforms \transfer by $6.8\%$-point on average and in particular $15.5\%$-points on CUB200, whereas the only loss in performance is on ChestX with a drop of $0.9\%$-point.
\begin{table*}[!t]
    \centering
    \input{tables/5_labeled_results.tex}

    \caption{Distillation on the eight target tasks with only 5\% labeled samples per task. Again, we compare to the baseline of \transfer{}. The largest value for each target task is in \textbf{bold}.}
    \label{tbl:5_labeled_results}
\end{table*}

\subsection{Different Measures of Correlation}
In order to evaluate the quality of a task similarity metric to estimate the performance of a target model after distillation, we consider the correlation between the computed metric and the actual observed performance after distillation. However, since we have no reason to believe that the relationship is linear, we consider the Spearman correlation in the main paper. However, for completeness of exposition, we report Pearson correlation and Kendall's Tau in Table \ref{tbl:single_teacher_pearson_corr} and Table \ref{tbl:single_teacher_kendall_corr}, respectively. For both these correlation measures, the overall conclusions are the same: Using feature representations is preferable to pseudo-labels, and PARC generally outperforms both CKA and RSA, albeit not by much over CKA.

\begin{table*}[htbp]
    \centering
    \input{tables/single_teacher_pearson_corr.tex}

    \caption{Pearson correlation between test accuracy after all possible single-source distillations and task similarity associated with the source models. Similar to Table \ref{tbl:single_teacher_spearman_corr}.}
    \label{tbl:single_teacher_pearson_corr}
\end{table*}

\begin{table*}[htbp]
    \centering
    \input{tables/single_teacher_kendall_corr.tex}

    \caption{Kendall Tau correlation between test accuracy after all possible single-source distillations and task similarity associated with the source models. Similar to Table \ref{tbl:single_teacher_spearman_corr}.}
    \label{tbl:single_teacher_kendall_corr}
\end{table*}

\subsection{Choice of Task Similarity Metrics}
Recently, multiple measures intended to estimate the transferability of a source model have been proposed. However, despite the very recently published Multi-Source Leep (MS-LEEP) and Ensemble Leep (E-Leep) no task similarity metric considers the estimation over multiple models at once \citep{agostinelli2022transferability}. Thus, we consider each source model separately and compute the metrics independent of other source models. This has the added benefit of reducing the number of metric computations required as we do not need to compute the task similarity for all possible combinations of $n$ models from $S$ possible (i.e. $\binom{n}{S}$), which grows fast with $S$.

Assume $\bX \in \R^{N \times d_X}$ and $\bY  \in \R^{N \times d_Y}$, and that $\bK_{ij} = k(\bx_i, \bx_j)$ for and $\bL_{ij} = l(\by_i, \by_j)$ where $k$, and $l$ are two (similarity) kernels as well as $\bx_i, \bx_j$ and $\by_i, \by_j$ are rows of $\bX$ and $\bY$, respectively. Then we have that CKA is defined as
\begin{align*}
    \rho_{\mathrm{CKA}}(\bX, \bY) \eqdef \frac{\mathrm{HSIC}(\bK, \bL)}{\sqrt{\mathrm{HSIC}(\bK, \bK)\mathrm{HSIC}(\bL, \bL)}},
\end{align*}
where $\bK, \bL \in \R^{N \times N}$ and HSIC is the Hilbert-Schmidt Independence Criterion,
\begin{align*}
    \mathrm{HSIC}(\bK, \bL) &\eqdef \frac{1}{(N-1)^2} \mathrm{tr}\left(\bK \bH_N \bL \bH_N \right), \quad \text{with } \\
    \bH_N &\eqdef \bI_N - \frac{1}{N}\mathbf{1}\mathbf{1}^\trans.
\end{align*}
In particular, if both $k$ and $l$ are linear kernels, then
\begin{align*}
    \rho_{\mathrm{CKA}}(\bX, \bY) = \frac{\lVert \bY^\trans \bX \rVert_F^2}{\lVert \bX^\trans \bX \rVert_F \lVert \bY^\trans \bY \rVert_F},
\end{align*}
where $\lVert \cdot \rVert_F$ is the Frobenius norm. We use the linear kernel throughout this paper and refer to \citet{Cortes2012AlgorithmsAlignment} for additional details on CKA.

For RSA, we consider the dissimilarity matrices given by
\begin{align*}
    \bK_{ij} &\eqdef 1 - \mathrm{pearson}(\bx_i, \bx_j) \quad \text{and} \\
    \bL_{ij} &\eqdef 1 - \mathrm{pearson}(\by_i, \by_j),
\end{align*}
where $\bX$ and $\bY$ are assumed normalized to have mean $0$ and variance $1$. We then compute RSA as the Spearman correlation between the lower triangles of $\bK$ and $\bL$, 
\begin{align*}
    \rho_{\mathrm{RSA}}(\bX, \bY) \eqdef \mathrm{spearman}\left(\{\bK_{ij} \mid i < j\}, \{\bL_{ij} \mid i < j\}\right).
\end{align*}
For additional details on RSA, we refer the reader to \citet{Dwivedi2019RepresentationLearning}. While \citet{Bolya2021ScalableLearning} introduces PARC alongside a heuristic and feature reduction, the PARC metric is almost identical to RSA. However, RSA was introduced to compute similarities between two sets of representations, and PARC was aimed at computing similarities between a set of representations and a set of labels associated with the dataset. Thus, in our use of PARC, it merely differs from RSA in the lack of normalization of $\bY$, which is assumed to be one-hot encoded vectors of class labels from the probe dataset.

\section{Experimental Details}\label{app:experiment_details}
In the following, we provide some experimental details.

\subsection{Main Experiments}
Unless otherwise mentioned, we use SGD with a learning rate of $0.01$, weight decay of $0.0001$, batch size of $128$, and loss weighting of $\lambda = 0.8$.
We initialize our target models with the ImageNet pre-trained weights available in torchvision (\url{https://pytorch.org/vision/stable/models}) and consider 28 fine-tuned models from \citet{Bolya2021ScalableLearning} publicly available at \url{github.com/dbolya/parc} as our set of source models. The source models consist of each of the architectures (AlexNet, GoogLeNet, ResNet-18, and ResNet-50) trained on CIFAR-10, Caltech101, CUB200, NABird, Oxford Pets, Stanford Dogs, and VOC2007. Note, we always exclude any source model trained on the particular target task, thus effectively reducing the number of source models for some target tasks.
For FixMatch we use a batch size of $128$ (with a 1:1 ratio of labeled to unlabeled samples for each batch) and fix the confidence threshold at $0.95$ and the temperature at $1$. We keep the loss weighting between the supervised loss and the unlabeled FixMatch loss at $\lambda = 0.8$.

\subsection{VTAB Experiments}
For each VTAB experiment, we consider the full training set (as introduced in \citet{Zhai2019ABenchmark}) as the unlabeled set, $\D_{\tau}^{u}$, and the VTAB-1K subset as the labeled set, $\D_{\tau}^{l}$. We use the Pytorch implementation from \citet{Jia2022VisualTuning} available at \url{github.com/KMnP/vpt}.

We use SGD with a learning rate of $0.005$, weight decay of $0.0001$, batch size of $128$ equally split in $64$ labeled and unlabeled samples, and loss weighting of $\lambda = 0.9$. We train our models for $100$ epochs, where we define one epoch as the number of steps required to traverse the set of unlabeled target data, $\D_{\tau}^{u}$ when using semi-supervised methods, or merely as the number of steps to traverse the labeled set, $\D_{\tau}^{l}$, for supervised transfer methods.
We initialize our target models with the BiT-M ResNet-50x1 model fine-tuned on ILSVRC-2012 from BiT \citep{Kolesnikov2020BigLearning} publicly available at \url{github.com/google-research/big_transfer}.

We consider the 19 BiT-M ResNet-50x1 models fine-tuned on the VTAB-1K target tasks from \citet{Kolesnikov2020BigLearning} as the set of source models. We always exclude the source model associated with the target task from the set of source models, and thus effectively have 18 source models available for each target task in VTAB. We use the PARC metric on the source model features to compute the source weighting, but also only use the top-$5$ highest-ranked source models to reduce the computational costs of training. Furthermore, we use $p=9$ for \multi{}.

\section{Domain gap between source tasks, targets tasks and ImageNet}\label{app:source_data}
As is evident from Figure \ref{fig:natural_images_overall} and Table \ref{tbl:vs_finetuned_source_models}, both \single{} and \multi{} do not yield notable improvements on e.g. ChestX and ISIC, but yield significant improvements on e.g. CUB200 and Oxford Pets. Notably, for the latter target tasks there are semantically similar source tasks present in our set of source models, while this is not true for the former target tasks. Hence, as one would expect, the availability of a source model trained on source tasks similar to the target tasks is important for cross-domain distillation to work well, which is expected to be true for both \single{} and \multi{}. Indeed, the task similarity metrics considered in this paper all aim at measuring alignment between tasks, and if the alignment between source and target tasks is small, we do not expect to gain much from distillation. This is affirmed by our experiments in \eg Table \ref{tbl:vs_finetuned_source_models}.

\subsection{A note on potential data overlap between source and target tasks}
Whenever any type of transfer learning is applied, including using ImageNet initializations, we (often implicitly) assume that the model we transfer from has not been trained on any data from the target test set. Although this assumption is often satisfied in practice due to domain gaps between the source and target task, utilizing initializations trained on e.g. ImageNet can potentially violate the assumption. This is due to the fact that ImageNet and many other modern publicly available datasets are gathered from various public websites and overlaps between samples in different datasets might occur.

Thus, it is natural to question whether the observed improvements are due to methodological advances or information leakage between source and target tasks. To ensure our advancements are valid we carefully remove any source model associated with the target task from the set of source models, $\S$. However, information leakage might still appear if \eg there are identical samples in the target dataset and the source dataset or ImageNet.
Despite large overlaps being improbable, it has been shown that there e.g. is a minor overlap (of at least 43 samples) between the training set of ImageNet and the test set of CUB200 (see e.g. \url{https://gist.github.com/arunmallya/a6889f151483dcb348fa70523cb4f578}). However, since the test set of CUB200 consists of 5794 samples, the presence of such a minor overlap should not affect the true performance of a model much.

In our experiments, we consistently compare our target models (initialized with ImageNet weights) to either identically initialized target models or source models initialized with either ImageNet weights or with weights from a source task. Hence, any potential gain from information leakage between ImageNet and a target task would bias both our results and the baselines, and thereby not affect our overall results. Furthermore, while an overlap between a source and target task might unfairly benefit the performance of our methods compared to \transfer{} and \fixmatch{}, such an overlap would likely benefit the fine-tuned source models even more making this baseline even harder to outperform (see e.g. Figure \ref{fig:acc_vs_compute} and Table \ref{tbl:vs_finetuned_source_models}).
Thus, our results should be at most as biased as the baselines.

%% file: tables/vtab/full.tex
\fontsize{6.5pt}{6.5pt}\selectfont
\newcolumntype{C}{>{\leavevmode\color{lightgray}\centering\arraybackslash}X}
\newcolumntype{M}{>{\leavevmode\color{black}\centering\arraybackslash}X}
\setlength{\tabcolsep}{0pt}
\setlength{\extrarowheight}{5pt}
\renewcommand{\arraystretch}{0.75}
\begin{tabularx}{\linewidth} {p{2.0cm}!{\color{lightgray}\vline} CCCCCCC!{\color{lightgray}\vline}M!{\color{lightgray}\vline}CCCC!{\color{lightgray}\vline}M!{\color{lightgray}\vline}CCCCCCCC!{\color{lightgray}\vline}M!{\color{lightgray}\vline}M}
\toprule
 & \rotatebox{90}{\raisebox{0.5pt}{\tikz\fill[natural] (0,0) circle (.5ex);} Caltech101} & \rotatebox{90}{\raisebox{0.5pt}{\tikz\fill[natural] (0,0) circle (.5ex);} CIFAR-100} & \rotatebox{90}{\raisebox{0.5pt}{\tikz\fill[natural] (0,0) circle (.5ex);} DTD} & \rotatebox{90}{\raisebox{0.5pt}{\tikz\fill[natural] (0,0) circle (.5ex);} Flowers102} & \rotatebox{90}{\raisebox{0.5pt}{\tikz\fill[natural] (0,0) circle (.5ex);} Pets} & \rotatebox{90}{\raisebox{0.5pt}{\tikz\fill[natural] (0,0) circle (.5ex);} SVHN} & \rotatebox{90}{\raisebox{0.5pt}{\tikz\fill[natural] (0,0) circle (.5ex);} Sun397} & \rotatebox{90}{\raisebox{0.5pt}{\tikz\fill[natural] (0,0) circle (.5ex);} Natural} & \rotatebox{90}{\raisebox{0.5pt}{\tikz\fill[specialized] (0,0) circle (.5ex);} Camelyon} & \rotatebox{90}{\raisebox{0.5pt}{\tikz\fill[specialized] (0,0) circle (.5ex);} EuroSAT} & \rotatebox{90}{\raisebox{0.5pt}{\tikz\fill[specialized] (0,0) circle (.5ex);} Resisc45} & \rotatebox{90}{\raisebox{0.5pt}{\tikz\fill[specialized] (0,0) circle (.5ex);} Retinopathy} & \rotatebox{90}{\raisebox{0.5pt}{\tikz\fill[specialized] (0,0) circle (.5ex);} Specialized} & \rotatebox{90}{\raisebox{0.5pt}{\tikz\fill[structured] (0,0) circle (.5ex);} Clevr-Count} & \rotatebox{90}{\raisebox{0.5pt}{\tikz\fill[structured] (0,0) circle (.5ex);} Clevr-Dist} & \rotatebox{90}{\raisebox{0.5pt}{\tikz\fill[structured] (0,0) circle (.5ex);} DMLab} & \rotatebox{90}{\raisebox{0.5pt}{\tikz\fill[structured] (0,0) circle (.5ex);} KITTI-Dist} & \rotatebox{90}{\raisebox{0.5pt}{\tikz\fill[structured] (0,0) circle (.5ex);} dSpr-Loc} & \rotatebox{90}{\raisebox{0.5pt}{\tikz\fill[structured] (0,0) circle (.5ex);} dSpr-Ori} & \rotatebox{90}{\raisebox{0.5pt}{\tikz\fill[structured] (0,0) circle (.5ex);} sNORB-Azim} & \rotatebox{90}{\raisebox{0.5pt}{\tikz\fill[structured] (0,0) circle (.5ex);} sNORB-Elev} & \rotatebox{90}{\raisebox{0.5pt}{\tikz\fill[structured] (0,0) circle (.5ex);} Structured} & \rotatebox{90}{\raisebox{0.5pt}{\tikz\fill[all] (0,0) circle (.5ex);} Mean} \\
\midrule
IN+Transfer & 88.1 & 47.0 & 57.4 & 85.8 & 82.8 & 75.3 & 27.8 & 66.3 & \textbf{81.0} & 95.0 & 80.0 & \textbf{72.7} & 82.2 & \textbf{73.1} & \textbf{55.9} & 43.6 & \textbf{75.7} & 18.7 & \textbf{58.6} & 21.2 & \textbf{46.0} & \textbf{49.1} & \textbf{62.4} \\
\multi{} & 88.6 & \textbf{60.9} & \textbf{62.4} & 86.1 & 84.4 & 79.0 & \textbf{38.4} & \textbf{71.4} & 80.6 & \textbf{95.9} & \textbf{83.3} & 72.2 & \textbf{83.0} & 57.4 & 45.6 & 44.6 & 67.7 & \textbf{27.4} & 44.9 & 23.9 & 38.2 & 43.7 & 62.2 \\
\single{} & \textbf{88.9} & 59.5 & 61.9 & \textbf{86.2} & \textbf{84.5} & \textbf{79.5} & 37.6 & 71.1 & 80.5 & 95.8 & 83.2 & 71.7 & 82.8 & 60.5 & 45.4 & \textbf{45.2} & 67.9 & 20.8 & 40.6 & \textbf{24.2} & 36.5 & 42.6 & 61.6 \\
\bottomrule
\end{tabularx}

%% file: tables/single_teacher_relative_acc_top_1.tex
\fontsize{7pt}{7pt}\selectfont
\newcolumntype{C}{>{\centering\arraybackslash}X}
\setlength{\tabcolsep}{0pt}
\setlength{\extrarowheight}{7pt}
\renewcommand{\arraystretch}{0.75}
\begin{tabularx}{1\linewidth}{p{0.4cm}p{0.9cm}!{\color{lightgray}\vline} CCCCCCCC !{\color{lightgray}\vline} C}
\toprule
 &  & \rotatebox{90}{CIFAR-10} & \rotatebox{90}{CUB200} & \rotatebox{90}{ChestX} & \rotatebox{90}{EuroSAT} & \rotatebox{90}{ISIC} & \rotatebox{90}{NABird} & \rotatebox{90}{Oxford Pets} & \rotatebox{90}{Stanford Dogs} & \rotatebox{90}{Mean} \\
\midrule
\multirow[c]{3}{*}{\rotatebox{90}{Pseudo}} & CKA & \textbf{99.6} & \textbf{100.0} & \textbf{96.1} & 99.5 & 98.1 & \textbf{100.0} & \textbf{100.0} & \textbf{100.0} & \textbf{99.2} \\
 & PARC & 99.3 & \textbf{100.0} & 93.6 & 99.5 & \textbf{98.3} & \textbf{100.0} & 98.4 & \textbf{100.0} & 98.6 \\
 & RSA & 99.3 & 74.8 & 94.8 & 99.5 & \textbf{98.3} & 86.6 & 97.8 & 95.6 & 93.4 \\
\midrule
\multirow[c]{3}{*}{\rotatebox{90}{Feature}} & CKA & \textbf{99.6} & 81.0 & 92.6 & \textbf{99.8} & \textbf{98.3} & \textbf{100.0} & \textbf{100.0} & \textbf{100.0} & 96.4 \\
 & PARC & \textbf{99.6} & \textbf{100.0} & 94.6 & 99.5 & 97.7 & \textbf{100.0} & 95.1 & \textbf{100.0} & 98.3 \\
 & RSA & \textbf{99.6} & \textbf{100.0} & 92.6 & 99.5 & \textbf{98.3} & 80.6 & \textbf{100.0} & \textbf{100.0} & 96.3 \\
\bottomrule
\end{tabularx}

%% file: tables/single_teacher_relative_acc_top_5.tex
\fontsize{7pt}{7pt}\selectfont
\newcolumntype{C}{>{\centering\arraybackslash}X}
\setlength{\tabcolsep}{0pt}
\setlength{\extrarowheight}{7pt}
\renewcommand{\arraystretch}{0.75}
\begin{tabularx}{1\linewidth}{p{0.4cm}p{0.9cm}!{\color{lightgray}\vline} CCCCCCCC !{\color{lightgray}\vline} C}
\toprule
 &  & \rotatebox{90}{CIFAR-10} & \rotatebox{90}{CUB200} & \rotatebox{90}{ChestX} & \rotatebox{90}{EuroSAT} & \rotatebox{90}{ISIC} & \rotatebox{90}{NABird} & \rotatebox{90}{Oxford Pets} & \rotatebox{90}{Stanford Dogs} & \rotatebox{90}{Mean} \\
\midrule
\multirow[c]{3}{*}{\rotatebox{90}{Pseudo}} & CKA & 99.3 & 98.7 & \textbf{98.3} & 99.7 & 99.0 & 92.9 & 99.2 & 98.4 & 98.2 \\
 & PARC & 99.7 & \textbf{100.0} & 96.7 & 99.7 & 98.9 & 94.5 & \textbf{99.4} & 98.4 & 98.4 \\
 & RSA & 99.7 & 83.2 & 97.6 & 99.8 & 99.0 & 84.9 & 99.2 & 92.8 & 94.5 \\
\midrule
\multirow[c]{3}{*}{\rotatebox{90}{Feature}} & CKA & \textbf{99.7} & 97.4 & 97.7 & 99.8 & 98.9 & 96.5 & 99.2 & 97.8 & 98.4 \\
 & PARC & \textbf{99.7} & 100.0 & 97.9 & 99.8 & 99.1 & \textbf{99.7} & 97.5 & \textbf{99.7} & \textbf{99.2} \\
 & RSA & \textbf{99.7} & 99.7 & 97.9 & \textbf{99.8} & \textbf{99.2} & 97.9 & 98.9 & \textbf{99.7} & 99.1 \\
\bottomrule
\end{tabularx}

%% file: tables/expert_metric_all_relative_acc.tex
\fontsize{7pt}{7pt}\selectfont
\newcolumntype{C}{>{\centering\arraybackslash}X}
\setlength{\tabcolsep}{0pt}
\setlength{\extrarowheight}{7pt}
\renewcommand{\arraystretch}{0.75}
\begin{tabularx}{0.75\linewidth}{p{1.0cm}!{\color{lightgray}\vline} CCC}
\toprule
 & CKA & PARC & RSA \\
\midrule
Pseudo & 0.985 & 0.990 & 0.974 \\
Feature & 0.986 & \textbf{0.993} & 0.991 \\
\bottomrule
\end{tabularx}

%% file: tables/general_results.tex
\fontsize{7pt}{7pt}\selectfont
\newcolumntype{C}{>{\centering\arraybackslash}X}
\setlength{\tabcolsep}{0pt}
\setlength{\extrarowheight}{5pt}
\renewcommand{\arraystretch}{0.75}
\begin{tabularx}{\linewidth}{p{2.6cm}!{\color{lightgray}\vline} CCCCCCCC !{\color{lightgray}\vline} C}
\toprule
 & \rotatebox{90}{CIFAR-10} & \rotatebox{90}{CUB200} & \rotatebox{90}{ChestX} & \rotatebox{90}{EuroSAT} & \rotatebox{90}{ISIC} & \rotatebox{90}{NABird} & \rotatebox{90}{Oxford Pets} & \rotatebox{90}{Stanford Dogs} & \rotatebox{90}{Mean} \\
\midrule
IN+Transfer & 92.4 & 42.8 & 47.3 & 97.4 & 81.6 & 37.3 & 75.9 & 62.6 & 67.2 \\
IN+FixMatch & \textbf{93.5} & 41.9 & 38.5 & \textbf{98.1} & \textbf{82.6} & 42.8 & 83.4 & 65.8 & 68.3 \\
\midrule
\multiEqual & 90.8 & 53.5 & 45.7 & 97.5 & 81.5 & 41.4 & 82.1 & 62.1 & 69.3 \\
\multi{1}  & 91.1 & 55.6 & 46.5 & 97.9 & 81.5 & 42.5 & 83.3 & 64.4 & 70.3 \\
\multi{3}  & 91.6 & 57.7 & 46.5 & 97.7 & 82.3 & 44.5 & 84.6 & 67.4 & 71.6 \\
\multi{6}  & 91.8 & 59.0 & 46.7 & 97.5 & 82.5 & 46.7 & \textbf{84.7} & 69.1 & 72.3 \\
\multi{9}  & 92.0 & 59.6 & 46.8 & 97.6 & 82.4 & 47.6 & 84.5 & 69.5 & 72.5 \\
\multi{12} & 92.0 & 60.0 & \textbf{47.7} & 97.6 & 82.2 & \textbf{48.3} & 84.4 & 69.9 & \textbf{72.8} \\
\multi{15} & 92.6 & \textbf{60.3} & 46.7 & 97.5 & 81.7 & 48.2 & 83.9 & 70.2 & 72.6 \\
\single{}  & 92.0 & 59.6 & 46.8 & 97.4 & 81.0 & 47.4 & 81.9 & \textbf{71.3} & 72.2 \\
\bottomrule
\end{tabularx}

%% file: tables/resnet50_vs_finetuned_source.tex
\fontsize{7pt}{7pt}\selectfont
\newcolumntype{C}{>{\centering\arraybackslash}X}
\setlength{\tabcolsep}{0pt}
\setlength{\extrarowheight}{5pt}
\renewcommand{\arraystretch}{0.75}
\begin{tabularx}{\linewidth}{p{2.5cm}!{\color{lightgray}\vline\hspace{0.2cm}}p{1.1cm}!{\color{lightgray}\vline} CCCCCCCC !{\color{lightgray}\vline} C}
\toprule
 & \shortstack[l]{Model\\Init.} & \rotatebox{90}{CIFAR-10} & \rotatebox{90}{CUB200} & \rotatebox{90}{ChestX} & \rotatebox{90}{EuroSAT} & \rotatebox{90}{ISIC} & \rotatebox{90}{NABird} & \rotatebox{90}{Oxford Pets} & \rotatebox{90}{Stanford Dogs} & \rotatebox{90}{Mean} \\
\midrule
IN+Transfer & ImageNet & 92.9 & 42.0 & 43.4 & 96.8 & 79.9 & 39.9 & 83.3 & 65.9 & 68.0 \\
Fine-tune Source & Source & \textbf{93.0} & \textbf{70.8} & 43.9 & 97.2 & \textbf{81.3} & 47.4 & 84.8 & \textbf{79.3} & \textbf{74.7} \\
\midrule
\multiEqual{} & ImageNet & 87.8 & 57.3 & 46.1 & 97.0 & 78.9 & 42.4 & 84.1 & 64.5 & 69.8 \\
\multi{12} & ImageNet & 91.5 & 64.5 & 45.4 & 97.0 & 78.9 & 49.8 & \textbf{87.1} & 74.2 & 73.6 \\
\midrule
\multiEqual{} & Source & 87.5 & 68.8 & 45.5 & \textbf{97.4} & 81.2 & 43.2 & 81.9 & 65.1 & 71.3 \\
\multi{12} & Source & 91.6 & 70.0 & \textbf{47.6} & 97.0 & 80.8 & \textbf{50.0} & 85.7 & 73.8 & 74.6 \\
\bottomrule
\end{tabularx}

%% file: tables/5_labeled_results.tex
\fontsize{7pt}{7pt}\selectfont
\newcolumntype{C}{>{\centering\arraybackslash}X}
\setlength{\tabcolsep}{0pt}
\setlength{\extrarowheight}{5pt}
\renewcommand{\arraystretch}{0.75}
\begin{tabularx}{\linewidth}{p{3.0cm}!{\color{lightgray}\vline} CCCCCCCC !{\color{lightgray}\vline} C}
\toprule
 & \rotatebox{90}{CIFAR-10} & \rotatebox{90}{CUB200} & \rotatebox{90}{ChestX} & \rotatebox{90}{EuroSAT} & \rotatebox{90}{ISIC} & \rotatebox{90}{NABird} & \rotatebox{90}{Oxford Pets} & \rotatebox{90}{Stanford Dogs} & \rotatebox{90}{Mean} \\
\midrule
IN+Transfer & 88.0 & 16.8 & \textbf{43.5} & 94.8 & 73.9 & 14.4 & 55.0 & 38.9 & 53.2 \\ 
\midrule
\multi{1} & 88.1 & 29.2 & 42.3 & \textbf{95.9} & 76.3 & 20.5 & 66.6 & 42.1 & 57.6 \\
\multi{9} & \textbf{90.2} & \textbf{32.3} & 42.6 & 95.9 & \textbf{76.7} & \textbf{24.8} & \textbf{68.2} & 49.0 & \textbf{60.0} \\
\single{} & 87.2 & 31.4 & 39.7 & 95.1 & 75.4 & 24.0 & 58.9 & \textbf{49.7} & 57.7 \\
\bottomrule
\end{tabularx}

%% file: tables/single_teacher_pearson_corr.tex
\fontsize{7pt}{7pt}\selectfont
\newcolumntype{C}{>{\centering\arraybackslash}X}
\setlength{\tabcolsep}{0pt}
\setlength{\extrarowheight}{7pt}
\renewcommand{\arraystretch}{0.75}
\begin{tabularx}{1\linewidth}{p{0.4cm}p{0.9cm}!{\color{lightgray}\vline} CCCCCCCC !{\color{lightgray}\vline} C}
\toprule
 &  & \rotatebox{90}{CIFAR-10} & \rotatebox{90}{CUB200} & \rotatebox{90}{ChestX} & \rotatebox{90}{EuroSAT} & \rotatebox{90}{ISIC} & \rotatebox{90}{NABird} & \rotatebox{90}{Oxford Pets} & \rotatebox{90}{Stanford Dogs} & \rotatebox{90}{Mean} \\
\midrule
\multirow[c]{3}{*}{\rotatebox{90}{Pseudo}} & CKA & 0.62 & \textbf{0.85} & 0.07 & 0.30 & -0.06 & 0.33 & 0.67 & 0.21 & 0.37 \\
 & PARC & 0.75 & 0.74 & -0.03 & 0.27 & -0.00 & 0.36 & 0.63 & 0.51 & 0.40 \\
 & RSA & 0.75 & 0.13 & -0.07 & 0.38 & 0.04 & -0.09 & 0.66 & 0.40 & 0.27 \\
\midrule
\multirow[c]{3}{*}{\rotatebox{90}{Feature}} & CKA & 0.84 & 0.60 & \textbf{0.39} & 0.29 & 0.00 & 0.30 & 0.71 & 0.54 & 0.46 \\
 & PARC & 0.86 & 0.73 & 0.17 & \textbf{0.46} & -0.06 & \textbf{0.58} & 0.77 & 0.78 & \textbf{0.54} \\
 & RSA & \textbf{0.90} & 0.85 & 0.07 & 0.45 & \textbf{0.04} & 0.27 & \textbf{0.87} & \textbf{0.83} & 0.54 \\
\bottomrule
\end{tabularx}

%% file: tables/single_teacher_kendall_corr.tex
\fontsize{7pt}{7pt}\selectfont
\newcolumntype{C}{>{\centering\arraybackslash}X}
\setlength{\tabcolsep}{0pt}
\setlength{\extrarowheight}{7pt}
\renewcommand{\arraystretch}{0.75}
\begin{tabularx}{1\linewidth}{p{0.4cm}p{0.9cm}!{\color{lightgray}\vline} CCCCCCCC !{\color{lightgray}\vline} C}
\toprule
 &  & \rotatebox{90}{CIFAR-10} & \rotatebox{90}{CUB200} & \rotatebox{90}{ChestX} & \rotatebox{90}{EuroSAT} & \rotatebox{90}{ISIC} & \rotatebox{90}{NABird} & \rotatebox{90}{Oxford Pets} & \rotatebox{90}{Stanford Dogs} & \rotatebox{90}{Mean} \\
\midrule
\multirow[c]{3}{*}{\rotatebox{90}{Pseudo}} & CKA & 0.51 & 0.46 & 0.16 & 0.28 & -0.05 & 0.24 & 0.49 & 0.07 & 0.27 \\
 & PARC & 0.61 & 0.64 & 0.01 & 0.12 & 0.02 & 0.36 & 0.54 & 0.39 & 0.34 \\
 & RSA & 0.62 & 0.17 & -0.07 & 0.22 & \textbf{0.08} & -0.01 & 0.48 & 0.29 & 0.22 \\
\midrule
\multirow[c]{3}{*}{\rotatebox{90}{Feature}} & CKA & 0.67 & 0.34 & \textbf{0.25} & 0.14 & -0.05 & 0.40 & 0.50 & 0.38 & 0.33 \\
 & PARC & 0.69 & \textbf{0.67} & 0.14 & \textbf{0.31} & -0.10 & \textbf{0.65} & 0.62 & 0.67 & \textbf{0.46} \\
 & RSA & \textbf{0.72} & 0.65 & 0.02 & 0.28 & 0.02 & 0.19 & \textbf{0.72} & \textbf{0.67} & 0.41 \\
\bottomrule
\end{tabularx}